\documentclass[format=acmsmall, screen=true]{acmart}
\AtBeginDocument{%
  }

\setcopyright{cc}
\setcctype{by}
\acmJournal{CSUR}
\acmYear{2026} \acmVolume{1} \acmNumber{1} \acmArticle{}
\acmMonth{1} \acmDOI{10.1145/3805796}

\usepackage{amsmath,amsfonts,bm}

\def\eqref#1{equation~\ref{#1}}

\def\floor#1{\lfloor #1 \rfloor}
\def\1{\bm{1}}

\def\va{{\bm{a}}}

\def\vh{{\bm{h}}}

\def\vm{{\bm{m}}}

\def\vp{{\bm{p}}}

\def\vs{{\bm{s}}}

\def\vx{{\bm{x}}}
\def\vy{{\bm{y}}}

\def\mA{{\bm{A}}}

\def\mC{{\bm{C}}}
\def\mD{{\bm{D}}}

\def\mH{{\bm{H}}}
\def\mI{{\bm{I}}}

\def\mP{{\bm{P}}}
\def\mQ{{\bm{Q}}}

\def\mW{{\bm{W}}}
\def\mX{{\bm{X}}}
\def\mY{{\bm{Y}}}

\DeclareMathAlphabet{\mathsfit}{\encodingdefault}{\sfdefault}{m}{sl}
\SetMathAlphabet{\mathsfit}{bold}{\encodingdefault}{\sfdefault}{bx}{n}

\def\gG{{\mathcal{G}}}

\newcommand{\E}{\mathbb{E}}

\newcommand{\R}{\mathbb{R}}

\DeclareMathOperator*{\argmax}{arg\,max}
\DeclareMathOperator*{\argmin}{arg\,min}

\def\V{{\mathcal{V}}}
\def\E{{\mathcal{E}}}
\def\A{{\mathcal{A}}}
\def\R{{\mathcal{R}}}

\def\N{{\mathcal{N}}}

\def\Lo{{\mathcal{L}}}

\usepackage{multirow}
\usepackage{enumitem}
\usepackage{makecell}
\usepackage{subfig}
\usepackage{algorithm}
\RequirePackage{algorithmic}
\usepackage{pifont}
\newcommand{\xmark}{\ding{55}}

\newcommand{\aggr}{\textsc{AGGR}}
\newcommand{\gnn}{\textsc{GNN}}
\newcommand{\update}{\textsc{UPDATE}}
\newcommand{\onehot}{\textsc{ONEHOT}}
\newcommand{\sample}{\textsc{SAMPLE}}

\begin{document}

\title{A Survey on Graph Neural Network Acceleration: Algorithms, Systems, and Customized Hardware}

\author{Shichang Zhang}
\email{shichang@cs.ucla.edu}
\affiliation{
  \institution{University of California, Los Angeles}
  \country{USA}
}

\author{Atefeh Sohrabizadeh}
\email{atefehsz@cs.ucla.edu}
\affiliation{
  \institution{University of California, Los Angeles}
  \country{USA}
}

\author{Cheng Wan}
\email{chwan@gatech.edu}
\affiliation{
  \institution{Georgia Institute of Technology}
  \country{USA}
}

\author{Zijie Huang}
\email{zijiehuang@cs.ucla.edu}
\affiliation{
  \institution{University of California, Los Angeles}
  \country{USA}
}

\author{Ziniu Hu}
\email{bull@cs.ucla.edu}
\affiliation{
  \institution{University of California, Los Angeles}
  \country{USA}
}

\author{Yewen Wang}
\email{wyw10804@cs.ucla.edu}
\affiliation{
  \institution{University of California, Los Angeles}
  \country{USA}
}

\author{Yingyan (Celine) Lin}
\email{celine.lin@gatech.edu}
\affiliation{
  \institution{Georgia Institute of Technology}
  \country{USA}
}

\author{Jason Cong}
\email{cong@cs.ucla.edu}
\affiliation{
  \institution{University of California, Los Angeles}
  \country{USA}
}

\author{Yizhou Sun}
\email{yzsun@cs.ucla.edu}
\affiliation{
  \institution{University of California, Los Angeles}
  \country{USA}
}

\renewcommand{\shortauthors}{Zhang et al.}

\begin{abstract}
Graph neural networks (GNNs) are emerging for machine learning research on graph-structured data. GNNs achieve state-of-the-art performance on many tasks, but they face scalability challenges when it comes to real-world applications that have numerous data and strict latency requirements. Many studies have been conducted on how to accelerate GNNs in an effort to address these challenges. These acceleration techniques touch on various aspects of the GNN pipeline, from smart training and inference algorithms to efficient systems and customized hardware. As the amount of research on GNN acceleration has grown rapidly, there lacks a systematic treatment to provide a unified view and address the complexity of relevant works. In this survey, we provide a taxonomy of GNN acceleration, review the existing approaches, and suggest future research directions. Our taxonomic treatment of GNN acceleration connects the existing works and sets the stage for further development in this area.

\end{abstract}

\ccsdesc[500]{Computing methodologies~Neural networks}
\ccsdesc[300]{Computer systems organization~Distributed architectures}
\ccsdesc[100]{Computer systems organization~Data flow architectures}
\ccsdesc[300]{Hardware~Integrated circuits}

\keywords{graph neural networks, model acceleration, hardware acceleration/accelerator}

\received{20 February 2007}
\received[revised]{12 March 2009}
\received[accepted]{5 June 2009}

\maketitle

\section{Introduction}\label{sec:introduction}
Graphs are a natural and powerful data structure for representing entities and their relationships. Many real-world data can be represented as graphs with nodes denoting entities and edges denoting their pairwise relationships, such as individuals in social networks, financial transactions, atoms and bonds in molecules, and vehicles in transportation systems. Graph neural networks (GNNs)~\cite{gcn, gat, sage} have become the most widely used graph machine learning (ML) model on graph data. GNNs have achieved state-of-the-art performance in many applications, including recommendations on social graphs \cite{zhang2019star, mao2021ultragcn, wu2020graph}, fraud detection on financial graphs \cite{fraud_detect}, drug discovery from molecule graphs~\cite{jiang2021could}, and traffic forecasting on transportation graphs~\cite{jiang2022graph}.

The superior performance of GNNs stems from combining entity information (node features) and relationships (graph structure). GNNs take node features and graph structure as input and output node representations for various tasks like node classification, edge prediction, and graph classification. For each node, a GNN learns its representation through \textit{message passing}, which aggregates messages (features) from the node's neighbors and updates the representation via a neural network transformation. Each aggregate-and-update operation is called a GNN layer, allowing the representation to combine information from direct neighbors. Stacking multiple GNN layers applies such operations recursively, enabling the representation to capture information from multi-hop neighbors and the neighborhood context. For a target node with an $L$-layer GNN, all neighbors within $L$ hops form its \textit{receptive field}. A recursively constructed $L$-layer tree graph with the target node as root and each node's neighbors as children is called a \textit{computation graph}.

As graph ML develops rapidly, graph data has become gigantic. For example, the Twitter user graph had 288M monthly active users (nodes) and an estimated 208 \textit{follow} relations (edges) per user as of 3/2015, while the Facebook user graph had 1.39B active users and over 400B total edges as of 12/2014 \cite{ching2015one}. Academic benchmarks have also grown from thousands of nodes, e.g., Cora~\cite{cora}, to as many as 240M nodes in the recent Open Graph Benchmark (OGB)~\cite{hu2021ogblsc}. Large graphs raise scalability challenges for both GNN training and inference. Unlike DNNs for images or text, where data can be processed via random mini-batch sampling due to the iid assumption, GNN nodes are interdependent. While GNNs leverage this dependency to learn informative representations, it also means multi-hop neighbors are involved in computation for every single node. This makes partitioning graph nodes into mini-batches nontrivial and causes computation graphs to grow exponentially with the number of GNN layers~\cite{oversquashing}. On large graphs where deeper GNNs are often needed~\cite{chenWHDL2020gcnii, li2021training}, both mini-batch training and single-node inference must process huge computation graphs. For instance, on the Twitter user graph with 208 neighbors per node, a three-layer GNN computation graph can contain millions of nodes (roughly $208^3$ minus overlaps). This node dependency problem makes applying GNNs on large graphs very challenging. Note that graph databases with many small graphs (e.g., chemical molecules with dozens to hundreds of nodes) do not face this issue and are not the focus of this survey.

Scalability challenges also affect GNN systems, including both Commercial-off-the-Shelf (COTS) systems and customized hardware. Large computation graphs require substantial memory, making adjacency matrix representations infeasible. In practice, graph structures must be processed in sparse matrix format. However, sparse matrices have different computation patterns from standard DNNs, preventing full exploitation of parallelism on devices like GPUs. This demands efficient systems with accelerated computation kernels and customized hardware accelerators like FPGAs.

\begin{figure*}[t]
\begin{center}
\includegraphics[width=0.9\textwidth]{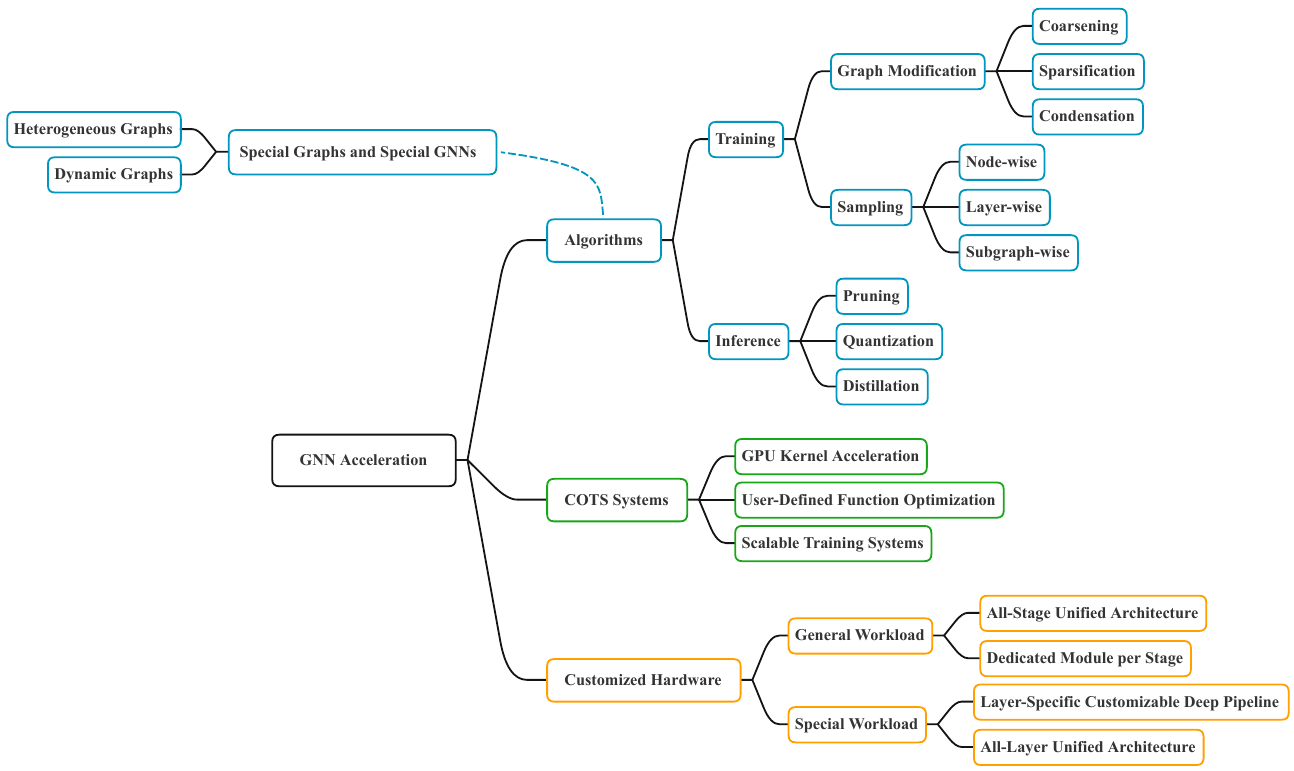}
\end{center}
\caption{A taxonomy of GNN acceleration. 
}
\Description{The figure illustrates a hierarchical taxonomy of GNN Acceleration, organized into three primary branches originating from a central root node labeled GNN Acceleration: Algorithms, COTS Systems, and Customized Hardware. The Algorithms branch is divided into Special Graphs and Special GNNs, Training, and Inference. Special Graphs and Special GNNs include Heterogeneous Graphs and Dynamic Graphs. The Training sub-branch includes Graph Modification (encompassing Coarsening, Sparsification, and Condensation) and Sampling (encompassing Node-wise, Layer-wise, and Subgraph-wise techniques). The Inference sub-branch includes Pruning, Quantization, and Distillation. The COTS Systems branch focuses on software and system-level optimizations, including GPU Kernel Acceleration, User-Defined Function Optimization, and Scalable Training Systems. The Customized Hardware branch is split into General Workload and Special Workload. General Workload includes All-Stage Unified Architecture and Dedicated Module per Stage. Special Workload includes Layer-Specific Customizable Deep Pipeline and All-Layer Unified Architecture.}
\label{figure:taxonomy}
\vskip -0.15in
\end{figure*}

The scalability challenges of GNNs require acceleration techniques for resource-efficient training and fast inference in latency-constrained applications. Acceleration techniques also benefit applications like neural architecture search. Many GNN acceleration techniques with different emphases have been proposed. In this survey, we provide a unified view of these techniques. Figure \ref{figure:taxonomy} shows our taxonomy, categorizing GNN acceleration into three main areas: (1) algorithms, (2) COTS systems, and (3) customized hardware. The algorithms category includes \textit{training} acceleration techniques that modify or sample from the graph to alleviate the node dependency problem, and \textit{inference} acceleration techniques like pruning, quantization, and distillation that convert trained GNNs into simpler, faster models. The system category includes GPU kernel acceleration for sparse matrix operations, user-defined function optimization for better hardware adaptation, and scalable distributed training systems. The customized hardware category includes accelerators with varying properties regarding layer customization, parallelization levels, and sparsity support. We also discuss techniques for special heterogeneous and dynamic graphs. We detail all techniques regarding their advantages and limitations, and suggest future research directions. Appendix~\ref{sec:background} has our notation table and a review of GNNs.

\section{GNN Acceleration: Training Algorithms} \label{sec:method_algorithm_training}
Given a $\gnn_{\theta}$ with randomly initialized parameters $\theta$ and a graph $\gG$ with node features $\mX$ and labels $\mY$, the goal of GNN training is to find an optimal set of parameters $\theta^*$ that minimizes the loss, i.e., $\theta^* = \argmin_{\theta} \Lo (\gnn_{\theta}(\gG, \mX), \mY)$. The major latency for GNN training algorithms comes from aggregating messages from (multi-hop) neighbor nodes in the receptive field, and the computation graphs can become huge for deep GNNs on large graphs. The general idea of GNN training acceleration is thus to reduce the computation graphs. Also, training with acceleration is desired to produce a GNN with similar performance to the GNN trained without acceleration. 

In this section, we discuss two training acceleration methods: graph modification and sampling. Both methods aim to reduce the computation graph to accelerate training. The core difference is whether there is a modified graph $\gG'$ as an intermediate output or only implicitly generated during sampling. Graph modification has two steps. The first step outputs a graph $\gG'$ that is smaller than $\gG$ and can be fastly trained on. The second step is regular GNN training with $\gG'$. For sampling, a subset of nodes/edges is selected to construct smaller computation graphs in each training iteration. Generally speaking, sampling also modifies $\gG$, but dynamically and implicitly, so there is no intermediate $\gG'$ output. Since the computation graph can differ every iteration, all the nodes/edges have the chance to be covered by sampling. In contrast, not all the nodes/edges will appear in $\gG'$ for graph modification, and new nodes/edges may be created.

\subsection{Graph Modification} \label{subsec:graph_modification}
\begin{figure*}[t]
\begin{center}
\includegraphics[width=0.8\textwidth]{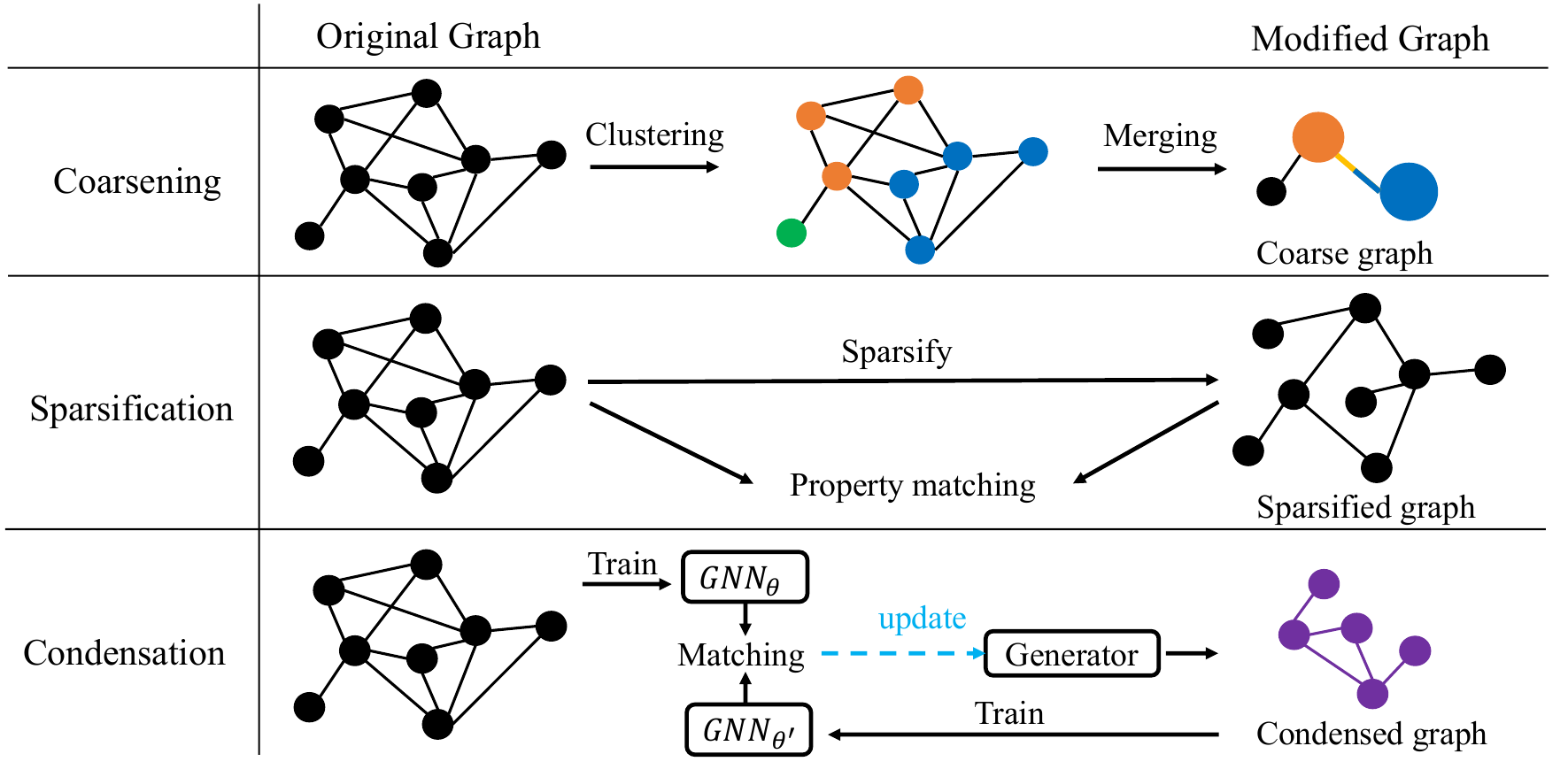}
\end{center}
\caption{\textbf{Illustration of graph modification methods:} \textit{Graph Coarsening} performs clustering and merge clusters of nodes into a super-node. \textit{Graph Sparsification} removes less important edges. \textit{Graph Condensation} generates a new condensed graph using a randomly initialized generative model. For the modified graph, black nodes/edges are from the original graph, and colored nodes/edges are newly created.}
\Description{The figure illustrates three distinct workflows for Graph Modification: Coarsening, Sparsification, and Condensation. Each begins with an Original Graph and results in a Modified Graph. Coarsening uses clustering and merging. Sparsification uses property matching to create a sparsified graph. Condensation uses a generator and a matching loop to create a small condensed graph.}
\label{figure:modification}
\vskip -0.2in
\end{figure*}

Graph modification accelerates GNN training in two steps. The first step takes the graph $\gG = (\V, \E)$, node features $\mX$, and labels $\mY$ as input and outputs modified $\gG' = (\V', \E')$, $\mX'$, and $\mY'$ for training. Training on $\gG'$ will be faster, if $\gG'$ is smaller than $\gG$ in the sense that $|\V'| < |\V|$ and/or $|\E'| < |\E|$. The second step trains a GNN on $\gG'$ such that the GNN has a similar performance on $\gG$ as if it is trained on $\gG$. Mathematically, regular training outputs $\theta^* = \argmin_{\theta} \Lo (\gnn_{\theta}(\gG, \mX), \mY)$, and training with graph modification outputs $\theta'^* = \argmin_{\theta'} \Lo (\gnn_{\theta'}(\gG', \mX'), \mY')$ such that $\gnn_{\theta'^*}$ has similar performance as $\gnn_{\theta^*}$ on $\gG$. In the following, we go over several types of graph modification: coarsening, sparsification, and condensation. 
Each of these methods modifies the graph $\gG$ differently focusing on different graph characteristics, but all resulting $\gG'$ are smaller graphs that accelerate GNN training. An illustration of graph modification methods is in Figure \ref{figure:modification}.

\subsubsection{\textbf{Graph Coarsening}} Graph coarsening aims to reduce the graph size while preserving its overall structure. Given a graph $\gG$, coarsening merges nodes in the same local structure into a single ``super-node'' and edges connecting super-nodes into a ``super-edge''. The coarse graph $\gG'$ thus has fewer nodes and edges compared to $\gG$. The key step of coarsening is to cluster nodes in $\gG$ into $K$ clusters for merging. Nodes in the same cluster will be merged into a super-node, and then a super-edge $(k, o)$ connecting super-node $k$ and $o$ will be constructed if there are edges connecting nodes in cluster $k$ and $o$. An illustration of the clustering and merging is in Figure \ref{figure:modification} top row. 

Graph coarsening, with its central step being clustering, has been studied long before GNNs were proposed~\cite{kernighan1970efficient, metis, spectral_clustering}, and new coarsening algorithms have been proposed until very recently \cite{loukas_coarsening2018, loukas_coarsening2019, bravo2019unifying}. The key idea of these algorithms is to coarsen the graph while preserving some graph properties, which are often related to the graph spectrum, e.g., restricted spectral approximation~\cite{loukas_coarsening2018} and inverse Laplacian~\cite{bravo2019unifying}. Related work includes Huang et al.~\cite {scale_up} for a general framework of graph coarsening to accelerate GNN training. GOREN~\cite{cai2021graph} for GNNs as a tool to help do better graph coarsening, where the goal is to output a coarse graph with better Rayleigh loss and Eigenerror. GraphZoom~\cite{Deng2020GraphZoom} for using graph coarsening to improve the accuracy and scalability of graph embedding algorithms like DeepWalK~\cite{perozzi2014deepwalk} and Node2Vec~\cite{grover2016node2vec}. More details are in Appendix~\ref{app:alg_training}

\subsubsection{\textbf{Graph Sparsification}} \label{subsubsec:sparsification}
These algorithms remove redundant edges to reduce the computation graph and accelerate GNN training, as in Figure \ref{figure:modification}. Like coarsening, graph sparsification has been studied widely before GNNs became popular~\cite{benczur1996approximating, spielman2011graph, spielman2014nearly, serrano2009extracting}. These methods usually try to find a sparsified $\gG'$ that well approximates the properties of the original $\gG$ by keeping all the nodes and removing less-important edges without disconnecting $\gG$. Example properties include the total weight of cuts \cite{benczur1996approximating}, spectral properties \cite{spielman2011graph, spielman2014nearly}, and hierarchical structures \cite{serrano2009extracting}. 

To accelerate GNN training, the above graph sparsification algorithms can be applied as a pre-processing step before GNN training starts. Since the sparsification algorithms are usually highly scalable and only need to be run once, the time it takes is negligible compared to GNN training. A sparsified graph can reduce the full-batch time complexity of training an $L$-layer GNN from $O(L|\E|)$ to $O(L|\E'|)$, where $|\E'|$ could be much smaller than $|\E|$.

Some related works perform GNN-based graph sparsification to improve GNN accuracy~\cite{li2020sgcn} or robustness~\cite{zheng2020robust} but not for GNN acceleration explicitly. Their difference from the traditional methods is the sparsification criteria consider both $\gG$ and the GNN instead of the properties of $\gG$ only. These criteria are usually formed as optimization problems and solved while training the GNN. For example, \cite{li2020sgcn} finds a sparsified graph $\gG'$ via the alternating direction method of multipliers. \cite{zheng2020robust} learns a sparsification strategy by generating sparse k-neighbor subgraphs. It is not guaranteed that these sparsification methods will always accelerate single GNN training. Although the sparse graph reduces $\E$ to accelerate message passing and is likely to make the model training converge faster, solving the extra optimization problem takes time and can slow down the training. These methods may be used to accelerate the training of the GNNs after the first one if multiple GNNs need to be trained on the same graph, as for the neural architecture search task. 

Pruning is a related concept sometimes confused with sparsification, though they differ: pruning typically refers to modifying the NN model, while sparsification modifies the graph. We discuss model pruning for GNN inference acceleration in Section \ref{subsec:pruning}. Graph pruning is broader than sparsification, as it can also involve node removal. However, in practice, removing entire nodes is rare, a more common approach is to dynamically exclude nodes per training iteration—a process better characterized as sampling, which is discussed in Section \ref{subsec:sampling}.

\subsubsection{\textbf{Graph Condensation}} Graph condensation generates a new graph that preserves the training dynamics of the original graph~\cite{jin2021graph, jin2022condensing}. In other words, the result of training two GNNs with identical architecture on the original graph $\gG$ and the condensed graph $\gG'$ (the generated new graph) should match. Meanwhile, $\gG'$ can be much smaller than $\gG$ thus accelerating GNN training. An illustration of graph condensation is shown in Figure \ref{figure:modification} bottom row.

\textbf{GCond}~\cite{jin2021graph} matches the training gradients of two GNNs to generate a condensed graph. In particular, a randomly initialized generative model first generates the condensed graph $\gG' = (\V', \E', \mX')$ and labels $\mY'$, given a condensation ratio $r$, and two GNNs $GNN_{\theta}$ and $GNN_{\theta'}$ are trained in parallel using stochastic gradient descent (SGD) on $\gG$ and $\gG'$ respectively. At each training step, SGD generates gradients $\nabla_{\theta} \Lo (GNN_{\theta}(\gG), \mY)$ and $\nabla_{\theta'} \Lo (GNN_{\theta'}(\gG'), \mY')$ for updating $\theta$ and $\theta'$ respectively. Then gradient matching is performed by tuning the generative model to minimize the distance between $\nabla_{\theta} \Lo (GNN_{\theta}(\gG), \mY)$ and $\nabla_{\theta'} \Lo (GNN_{\theta'}(\gG'), \mY')$, so that $\gG'$ and $\mY'$ better preserve the training dynamics can be generated in the next iteration. For the generative model, since $\E'$ and $\mX'$ depend on each other in general graphs, GCond chooses to generate $\mX'$ first and then construct $\E'$ from $\mX'$. Specifically, GCond initializes $r|\V|$ node features $\mX'$ as free learnable parameters, which will be directly updated by the gradient-matching loss. To generate condensed edges $\E'$, GCond uses an MLP as a feature-based structure prediction model, which takes input $\vx'_i$ and $\vx'_j$ and predicts edge $i$-$j$. Finally, for condensed labels $\mY'$, GCond initializes a $\vy'$ for each $\vx'_i$ and fixes $\vy'$ by randomly selecting a class following the label distribution in $\gG$ without learning. Since all $\mX'$ are randomly initialized, not the value of $\vy'$ assigned to each $\vx'_i$ but only the distribution of $\mY'$ matters. GCond sets the distribution $\mY'$ to follow the original $\mY$ so that the class ratios in $\gG$ are kept in $\gG'$. 

Note that like the GNN-based sparsification methods, GCond can only accelerate training for the second and subsequent GNNs when multiple GNNs need to be trained on the same graph because the gradient matching step requires training a GNN on the original graph $\gG$. However, GCond is shown to achieve an astonishing less than 1\% condensation ratio on several graph benchmarks while maintaining over 95\% accuracy as the original GNN, and the same condensed $\gG'$ can be transferred to train GNNs with different architectures to achieve good performance. It is thus a great fit for tasks like neural architecture search. A neural architecture search experiment is performed in GCond, and hundreds of GNNs were quickly trained to achieve competitive performance after a good $\gG'$ is generated. However, to accelerate the training of a single GNN, GCond should either be combined with other acceleration methods or better condensation strategies that can match $GNN_{\theta}$ and $GNN_{\theta'}$ more efficiently without training to convergence is needed.

\subsection{Sampling} \label{subsec:sampling}
Sampling is a prevalent technique for accelerating GNN training~\cite{liu2021sampling, serafini2021scalable}. The idea is to dynamically select a subset of nodes and edges to build the computation graph, greatly reducing its size and accelerating training. Model accuracy can often be maintained since the full neighborhood is not always required to predict a target node. The shared goal of different sampling algorithms is to reduce time and memory consumption while maintaining accuracy. Unlike graph modification, sampling is done dynamically without an intermediate modified graph output.

\begin{figure*}[t]
\begin{center}
\includegraphics[width=0.9\textwidth]{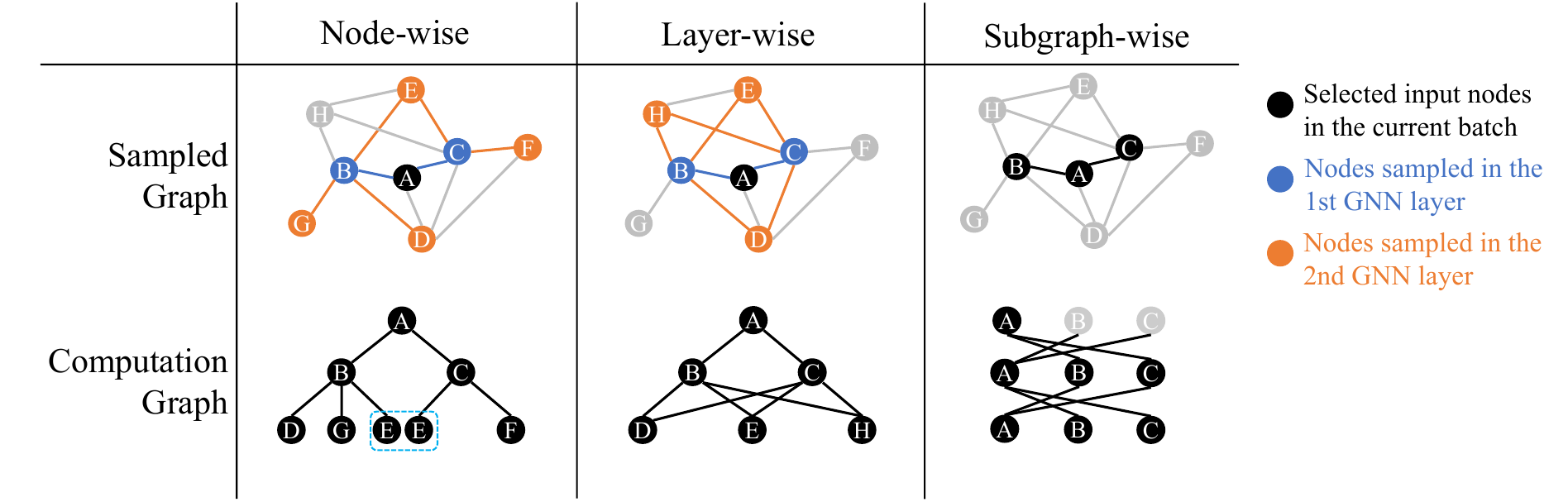}
\end{center}
\caption{\textbf{Illustration of different Graph Sampling Methods.} \textit{Node-wise Sampling} methods sample per node for each computation layer, which might lead to redundant nodes (e.g., node \textsc{E} is sampled twice), and missing edges (e.g., edge between node \textsc{C} and node \textsc{D} is missing); \textit{Layer-wise Sampling} methods sample per layer based on nodes in the previous layer. \textit{Subgraph-wise Sampling} methods sample a list of nodes and their induced subgraph, and then conduct message passing via all the edges within the sampled subgraph.}
\vskip -0.15in
\Description{The figure compares three sampling techniques: Node-wise, Layer-wise, and Subgraph-wise. Node-wise sampling prunes neighbors independently for each node. Layer-wise sampling selects a collective set of nodes per layer to reduce redundancy. Subgraph-wise sampling identifies a small induced subgraph and restricts all computations to that local structure. A color-coded legend distinguishes between input nodes, 1st-layer sampled nodes, and 2nd-layer sampled nodes.}
\label{figure:sampling}
\end{figure*}

We introduce different sampling algorithms in a unified framework. Throughout this section, we use $\sample(\cdot)$ to denote the sampling operation and $\overline{\cdot}$ to denote the sampled version. Given each target node $v$, instead of aggregating from all neighbors $\mathcal{N}(v)$ for message passing, sampling selects a subset $\overline{\mathcal{N}}(v)$ that provides the most important information for predicting $v$. 
A sampled version of GNN message passing can approximate the node representations, where $\overline{\N}(v)^{(l)}$ and $\overline{\mP}^{(l)}$ indicate the layer-$l$-sampled neighborhood and its filter matrix:
\begin{align} 
\vh_{v}^{(l)} &\approx \update (\vh_{v}^{(l-1)}, \aggr (\{\phi(\vh_v^{(l-1)}, \vh_u^{(l-1)}) | u \in \overline{\N}(v)^{(l)} \})), \label{eq:message_passing_sample}
\\
\mH^{(l)} &\approx \sigma (\overline{\mP}^{(l)} \mH^{(l - 1)} \mW^{(l)}). \label{eq:message_passing_matrix_sampled}
\end{align} 
The key idea of sampling is to find informative neighborhoods that construct small but effective computation graphs. Existing sampling algorithms can be categorized into three types according to their input to $\sample(\cdot)$: 1) \textbf{node-wise}, which samples within a single node neighborhood; 2) \textbf{layer-wise}, which jointly samples neighborhoods for all nodes in the same computation graph layer; 3) \textbf{subgraph-wise}, which samples a subgraph from the whole graph as neighborhoods for nodes.
Figure \ref{figure:sampling} shows the workflow of three sampling methods. 
We compare runtime and accuracy of popular sampling methods in Table \ref{tab: sampling performance} on three benchmarks: Pubmed \cite{cora}, PPI \cite{zitnik2017predicting}, and Reddit \cite{hamilton2017inductive}. Layer-wise methods work better on Reddit, and subgraph-wise methods work better on PPI.

\subsubsection{\textbf{Node-wise Sampling}} \label{subsubsec:node-wise}
These methods apply to each target node $v$ at each layer $l$. The sampled neighborhood is obtained by: $\overline{\N}(v)^{(l)}\gets\sample^{(l)}\big(\mathcal{N}(v)\big)$. Given $s_{node}^{(l)}$ as the number of neighbors to sample for each $v$ at layer $l$, the sampling operation $\sample^{(l)}(\mathcal{N}(v))$ is performed $s_{node}^{(l)}$ times. We use $u\sim p^{(l)}_v(u)$ to denote the probability node $u$ is sampled as part of the computation graph for node $v$. An example is in Figure~\ref{figure:sampling}(a).

\textbf{GraphSAGE} \citep{sage} pioneers to reduce node receptive field with node-wise sampling. For each layer $l$, GraphSAGE uniformly samples a fixed number ($s_{node}^{(l)}$) of neighbors for each node. Each neighbor node $u$ in $\N(v)$ has equal probability $1/d(v)$ to be sampled, i.e. $p^{(l)}_v(u) = 1/d(v)$. The same procedure is repeated to construct a computation graph layer smaller than the full neighborhood.
GraphSAGE can also be formalized in matrix form, with $\mP^{(l)} \mH^{(l - 1)}$ denoting full-neighbor aggregation and $\overline{\mP}^{(l)} \mH^{(l - 1)}$ denoting aggregation with sampling. The sampled filter matrix $\overline{\mP}^{(l)}$ can be computed by an element-wise multiplication with a mask matrix $\mathcal{M} \in\mathbb{R}^{N\times N}$ as $\overline{\mP}^{(l)} = \mathcal{M}\circ \mP^{(l)}$, where $\mathcal{M}$ is:
\begin{equation}
  \mathcal{M}_{u,v} =
    \begin{cases}
      \frac{|\mathcal{N}(v)|}{s_{node}^{(l)}} & u \in \overline{\N}(v)^{(l)} \ \  \text{, where} \ \ \overline{\N}(v)^{(l)}\gets\texttt{SAMPLE}^{(l)}\big(\mathcal{N}(v)\big)\\
      0 & \text{otherwise}
    \end{cases}       
\end{equation}
$\overline{\mP}^{(l)} \mH^{(l - 1)}$ is an unbiased estimation of ground-truth $\mP^{(l)} \mH^{(l - 1)}$, i.e., $\mathbb{E} \big[\overline{\mP}^{(l)} \mH^{(l - 1)} \big] = \mP^{(l)} \mH^{(l - 1)}$.

Many other works explore different aspects of node-wise sampling. VR-GCN \citep{chen2017stochastic} uses historical embeddings to accelerate training convergence. PinSAGE Sampler \citep{ying2018graph} uses random walk frequency-based sampling distribution for recommender systems. 
MVS-GNN \citep{cong2020minimal} extends VR-GCN by explicitly reducing both embedding approximation variance and stochastic gradient variance.
GCN-BS \citep{liu2020bandit} formulates neighbor sampling as a multi-armed bandit problem, proposing a learnable sampler updated towards minimal sampling variance. BNS \citep{yao2021blocking} first conducts random node sampling, then uniformly selects a small portion of sampled neighbors to stochastically block their ongoing expansion. More details on node sampling are in Appendix~\ref{app:alg_training}.
Node-wise sampling reduces the receptive field to address scalability but has limitations:
\begin{itemize}
    \item Exponential neighbor growth as the GNN goes deeper, even though neighborhood size is reduced. As in Figure \ref{figure:sampling}(a), the computation graph grows by a factor of $s_{node}^L$ in each layer.
    \item A node can be repeatedly sampled since sampling procedures for each node in the same layer are independent. In Figure \ref{figure:sampling}(a), node \textsc{E} is sampled by both node \textsc{B} and node \textsc{C}, and message passing computation is repeated twice (marked by the blue dotted box).  
    \item Some edges between sampled nodes across layers are missing, wasting computation resources. In Figure \ref{figure:sampling}(a), \textsc{C} and \textsc{D} are connected in the original graph, but since \textsc{D} is sampled by \textsc{B} but not \textsc{C}, edge \textsc{(B, C)} is missing, though both nodes are in the computation graph.
\end{itemize}

\begin{table}[tb]
\scriptsize
\caption{Performance comparison of different sampling methods. Absolute performance is adopted from \cite{liu2021sampling}, and we add relative performance with GraphSAGE set as the benchmark (1$\times$), as it is the simplest and the most widely used sampling method. All results assume a 2-layer backbone GCN model with the sampling size and the optimizer the same as the original papers.}
\begin{tabular}{cc|cc|cc|cc}
\toprule
                               &             & \multicolumn{2}{c|}{Pubmed}                & \multicolumn{2}{c|}{PPI}                     & \multicolumn{2}{c}{Reddit}                   \\
                               &             & Accuracy (\%)        & Total Time (s)        & Accuracy (\%)        & Total Time (s)          & Accuracy (\%)        & Total Time (s)          \\ \midrule
No Sample                      & GCN         & 78.6 (0.97$\times$) & 24.8 (0.65$\times$) & N/A                 & N/A                    & N/A                 & N/A                    \\ \hline
\multirow{2}{*}{Node-wise}     & GraphSAGE   & 81.4 (1$\times$)    & 37.9 (1$\times$)    & 61.6 (1$\times$)    & 47.5 (1$\times$)      & 95.0 (1$\times$)    & 326.7 (1$\times$)     \\
                               & VR-GCN      & 81.4 (1$\times$)    & 5.9 (0.16$\times$)  & 97.5 (1.58$\times$) & 83.5 (1.76$\times$)   & 96.3 (1.01$\times$) & 390.6 (1.20$\times$)  \\ \hline
\multirow{3}{*}{Layer-wise}    & FastGCN     & 86.3 (1.06$\times$) & 13.7 (0.36$\times$) & N/A                 & N/A                    & 92.8 (0.98$\times$) & 294.5 (0.90$\times$)  \\
                               & AS-GCN      & 89.5 (1.10$\times$) & 39.1 (1.03$\times$) & N/A                 & N/A                    & 96.5 (1.02$\times$) & 1633.7 (5.00$\times$) \\
                               & LADIES      & 75.7 (0.93$\times$) & 6.4 (0.17$\times$)  & N/A                 & N/A                    & N/A                 & N/A                    \\ \hline
\multirow{2}{*}{Subgraph-wise} & Cluster-GCN & N/A                 & N/A                  & 95.8 (1.56$\times$) & 202.3 (4.26$\times$)  & 95.8 (1.01$\times$) & 981.8 (3.00$\times$)  \\
                               & GraphSAINT  & N/A                 & N/A                  & 98.7 (1.60$\times$) & 550.1 (11.58$\times$) & 96.7 (1.02$\times$) & 124.0 (0.38$\times$)  \\ \bottomrule
\end{tabular}
\vskip -0.1in
\label{tab: sampling performance}
\end{table}

\subsubsection{\textbf{Layer-wise Sampling}}
These methods are applied to all nodes in the same computation layer instead of a single node. When a set of nodes is selected as the prediction target, a computation graph is constructed with all these nodes in the first computation layer. Their direct neighbors belong to the next layer, their 2nd-hop neighbors to the layer after, and so on. Let $\V^{(l)}$ denote all nodes at the $l$-th layer. Layer-wise sampling chooses a subset from $\V$ for the next computation layer. For each $v \in \mathcal{V}^{(l)}$, its layer-sampled neighborhood is the intersection of the sampled node set and its original neighborhood: $\overline{\N}(v)^{(l)}\gets \texttt{SAMPLE}(\V)\cap\mathcal{N}(v)$. Let $s_{layer}^{(l)}$ be the number of neighbors to sample for layer $l$, the sampling operation $\sample^{(l)}(\V)$ is performed $s_{layer}^{(l)}$ times.
Finally, the sampled nodes at the current computation layer $l$ would be used to construct the neighborhood of the later computation layer $l+1$. An example is in Figure~\ref{figure:sampling}(b).

\textbf{FastGCN} \citep{chen2018fastgcn} proposes the first layer-wise importance sampling scheme to address scalability issues in node-wise sampling. For each computation layer, FastGCN samples nodes from the full node set $\V$ with node-degree-based probability: $p^{(l)}_{\mathcal{V}}(u) = \|\mP_{:,u}\|^2_2 / \sum_{u' \in \mathcal{V} }  \|\mP_{:,u'} \|^2_2$. It uses the same sampling distribution for different layers, i.e. $p^{(l)}_{\mathcal{V}}(u) = p_{\mathcal{V}}(u)$ for all $l$. The approximation of $\mP^{(l)} \mH^{(l - 1)}$ with $\overline{\mP}^{(l)} \mH^{(l - 1)}$ can be computed by constructing
$\overline{\mP}^{(l)}$, with the $(u, v)$ entry being
\begin{align}
    \overline{\mP}^{(l)}_{v,u} = 
\begin{cases}
    \frac{\mP_{v,u}^{(l)}}{s_{layer}^{(l)}p_{\mathcal{V}}(u)} & \text{if } v \text{ is sampled in layer ($l$) and } u\in\mathcal{N}(v) \\
    0,& \text{otherwise}
    \end{cases}
    \label{eq: P for Layer-wise sampling}
\end{align}

Despite efficiency, FastGCN suffers from connectivity challenges since sampling for each layer is independent. Nodes sampled in two consecutive layers may not be connected, preventing some nodes from passing/receiving information to the next layer.
This leads to a sparse $\overline{\mP}^{(l)}$ matrix and may cause large variance in approximated node representations.
More layer-wise sampling algorithms like AS-GCN \citep{huang2018adaptive} and LADIES \citep{zou2019layer} are discussed in Appendix~\ref{app:alg_training}. Layer-wise sampling methods solve the exponential neighborhood expansion limitation of node-wise methods and enjoy linear time and memory complexity in layer $L$. However, the connectivity problem persists: sampled nodes in a later layer can have limited or no neighbors from the previous layer. Thus, messages passed between layers for some sampled nodes can be insufficient, degrading model performance. 

\subsubsection{\textbf{Subgraph-wise Sampling}}
These methods take the whole graph as input and output a sampled subgraph, i.e., $\overline{\gG} \leftarrow \sample(\gG)$. GNN training uses the sampled subgraph to conduct full-batch message passing. The sampling procedure can happen before training starts as a pre-processing step, making them similar to graph modification methods. However, two key differences distinguish them. First, to overcome scalability challenges, the subgraphs are usually much smaller than the original, representing only local information instead of global information as in graph modification methods (Section \ref{subsec:graph_modification}). Second, directly training with such pre-processed small subgraphs often results in low accuracy. Thus, some stochasticity needs to be added back during training to improve accuracy, making the modification in these methods dynamic as discussed below.

\textbf{ClusterGCN} \citep{chiang2019cluster} utilizes graph clustering algorithms (i.e., METIS~\citep{metis}) to partition the input graph $\mathcal{G}$ into clusters, where nodes within each cluster form a densely connected subgraph and inter-cluster edges are minimized. ClusterGCN then conducts full-batch GNN training on each subgraph cluster to improve scalability. One limitation is that the clustering result is fixed, so inter-cluster edges will be missing throughout training. ClusterGCN addresses this by randomly choosing subgraphs in each training mini-batch and adding inter-cluster edges between them. This operation covers provides an unbiased estimate of full-neighborhood node representations.

Many other follow-up works conduct subgraph-wise sampling. GraphSAINT \citep{zeng2019graphsaint} first samples nodes then constructs subgraphs induced by the sampled nodes. Shadow-GNN \citep{zeng2021decoupling} samples different subgraphs for each target node via random walk or personalized page rank. RWT \citep{bai2021ripple} proposes a ripple walk sampler that randomly samples a small set of nodes, then iteratively samples from their neighbors until a prespecified budget is filled. Zeng et al.~\citep{zeng2021accurate} provides a parallelized version of existing frontier sampling that can sample subgraphs approximating the original graph w.r.t. various connectivity measures. Additionally, some layer-wise sampling methods can be used to construct subgraphs~\cite{hu2020heterogeneous}. More details on subgraph sampling methods are in Appendix~\ref{app:alg_training}.
The key advantage of subgraph-wise sampling is that they are independent of the GNN model and generated embeddings, making them suitable for parallel or pre-training. However, this independence can also be a limitation, as the sampling only considers graph structure without accounting for model training dynamics. Therefore, it remains an open question how to incorporate variance-reduction studies, e.g., VR-GCN and AS-GCN, into subgraph-wise sampling.

\section{GNN Acceleration: Inference Algorithms} \label{sec:method_algorithm_inference}
Given $\gG_{train}$ and $\gG_{test}$ as the training and test graphs, and $\gnn_{\theta^*}$ as the optimal GNN trained on $\gG_{train}$, GNN inference generates representations for nodes in $\gG_{test}$ and uses them to make predictions. The goal of GNN inference acceleration is to construct another $\tilde \gnn_{\tilde \theta}$ such that $\tilde \gnn_{\tilde \theta}$ inference on $\gG_{test}$ has similar accuracy as $\gnn_{\theta^*}$ but is faster. During the construction of $\tilde \gnn_{\tilde \theta}$, access to $\gG_{test}$ is not assumed, since fast inference is desired on \textbf{unseen} test graphs for real-world use cases. In contrast, access to $\gG_{train}$ is usually assumed. In practice, $\gG_{test}$ and $\gG_{train}$ can have overlap or be the same graph (the transductive setting). For generality, we assume they are different.

GNN inference acceleration differs substantially from GNN training acceleration. Training acceleration is more graph-centric and inference acceleration is more model-centric. For training, the GNN architecture is fixed and only the process for finding $\theta^*$ is accelerated. Methods focus on modifying or sampling the graph to reduce the computation graph. For inference, $\tilde \gnn_{\tilde \theta}$ can differ substantially from $\gnn_{\theta^*}$ in model architecture, parameter precision, and parameter values. Changes are made only on the model. The test graph $\gG_{test}$ is assumed unseen and thus untouched. Only when $\gG_{test}$ and $\gG_{train}$ overlap or are the same can some graph-centric training acceleration methods like graph sparsification also accelerate inference because the sparsification carries from $\gG_{train}$ to $\gG_{test}$. Even then, graph modification methods like coarsening and condensation do not fit because they can affect the node for inference and thus lose the prediction target. Sampling methods are also rarely used because usually only one forward computation is performed in inference rather than many times in training. Therefore, sampling only a random partial neighborhood generally cannot guarantee prediction accuracy or consistency.

Fast inference with $\tilde \gnn_{\tilde \theta}$ does not mean $\tilde \gnn_{\tilde \theta}$ can be constructed fast. Rather, the speed of $\tilde \gnn_{\tilde \theta}$ construction is often sacrificed to achieve inference acceleration. Nevertheless, once a fast-inferring $\tilde \gnn_{\tilde \theta}$ is constructed, it can be useful in practice, especially when construction is infrequent but inference latency is critical.
We discuss three kinds of methods for inference acceleration: pruning, quantization, and distillation. All three are widely used to accelerate general DNNs. The core difference is whether they develop a new faster model (distillation) or perform in-place modifications, on original model parameters (pruning) or floating-point numbers (quantization) respectively.
The works we discuss below adapt them on GNNs for graphs. The major challenge is how to preserve graph structure information and deal with latency caused by message aggregation.

\subsection{Pruning} \label{subsec:pruning}
Model pruning is a popular approach for accelerating neural network (NN) inference \cite{han_pruning}, with the core idea being removing unimportant parts of an NN while maintaining accuracy~\cite{reed1993pruning}. The most common method is weight magnitude pruning, which prunes connections between NN ``neurons" with small weights or equivalently reduces the L1 norm of NN weights. Like other inference acceleration methods, pruning provides a trade-off between model speed and accuracy. When more weights are removed, the model typically becomes faster but less accurate, and vice versa.

Pruning methods have been proposed for accelerating GNN inference~\cite{zhou2021accelerating, chen2021unified, prunegnn}, where PruneGNN~\cite{prunegnn} is an optimized algorithm-architecture framework that speeds up training as well. Note that GNN pruning targets the model and differs from graph pruning which targets the graph data, discussed in Section \ref{subsubsec:sparsification}. For GNN model pruning, selecting which GNN weights to prune is the central question and is nontrivial. It is challenging because GNN inference on a graph node depends on its local graph structure and neighboring nodes, which is more complicated than regular NN inference on data types with independent instances like image classification.

\textbf{Zhou et al.} \cite{zhou2021accelerating} propose to accelerate GNN inference with channel pruning, where channels refer to the dimensions of the node representations $\vh$, or equivalently the rows of the GNN weight matrix $\mW$. Given a GNN with weights $\mW$, Zhou et al. formulate the pruning problem as a LASSO regression problem for each GNN layer, with the objective:
\begin{equation} \label{eq:channel_prune}
\argmin_{\hat{\bm{\beta}}, \hat\mW} \lVert \mH^{(l)} - \mP \mH^{(l - 1)} (\hat{\bm{\beta}} \odot \hat\mW^{(l)}) \rVert_{2} + \lambda \lVert \hat{\bm{\beta}} \rVert_{1}.
\end{equation} 
There are two sets of parameters to optimize: a learnable mask $\hat{\bm{\beta}} \in \R^F$ that chooses channels to prune from $F$ total channels, and the GNN weight matrix $\hat\mW$. $\hat{\bm{\beta}}$ is initialized as all ones and $\hat\mW$ is initialized as $\mW$. $\odot$ denotes element-wise product on each column. $\mP$ is the filter matrix. Zhou et al. conducts alternating optimization. $\hat{\bm{\beta}}$ is optimized first with fixed $\hat\mW$ and SGD. Then $\hat\mW$ can be computed in closed-form for fixed $\hat{\bm{\beta}}$. The L1 regularization shrinks entries of $\hat{\bm{\beta}}$ for pruning. 
Another line of pruning discussed in Appendix~\ref{app:alg_inference} is the Lottery Ticket Hypothesis (LTH) \cite{frankle2018lottery, chen2021unified}, which says that a pruned NN can be \textit{retrained} to achieve similar accuracy as the original NN.

\subsection{Quantization}
Quantization is a widely used acceleration technique. The idea is to use lower numerical precision for model parameters, say replace 32-bit floating-point numbers (FP32) with 8-bit integers (INT8). For DNNs, a significant part of latency comes from matrix multiplications, which boil down to multiply-accumulate (MAC) operations. Quantized low-precision DNNs have fewer MAC operations and reduce memory access time~\cite{hubara2018quantization, gupta2015deep}.  
The concern is that reducing the numerical precision can result in large accuracy loss. Thus, quantization methods aim for acceleration while maintaining accuracy \cite{krishnamoorthi2018quantizing, han2015deep}. Below, we first introduce quantization on general NNs and then GNN quantization.

Quantization on NNs boils down to operations on tensors, which could be weights or activations. Suppose we want to quantize a given tensor $\vx$ into a $q$-bit representation. Let $q_{min}$ and $q_{max}$ be the minimum and maximum values for the $q$-bit representations. Let $s$ be a factor that scales $\vx$ to the range $[q_{min}, q_{max}]$. Let $z$ be a special \textit{zero-point} meant to ensure the point zero is quantized with no error, which is important as NNs often have operations like zero padding. Both $s$ and $z$ are in $q$-bit. We use the floor function ($\floor{\cdot}$) to indicate cutting the representation to $q$-bit (i.e., quantizing to integers). Then the quantized tensor $\vx_q$ is $\vx_q = \floor{\frac{\vx}{s} + z}$.
The quantized $\vx_q$ is in the desired precision. A \textit{dequantization} is then performed to map $\vx_q$ to $\hat\vx$ as $\vx \approx \hat \vx = (\vx_q - z) s$.
With the (de)quantize operations, a $q$-bit representation $\hat \vx$ can replace $\vx$ for faster computation. 

With the same quantization idea, there are two types of algorithms for NNs: Post-Training Quantization (PTQ) and Quantization-Aware Training (QAT) \cite{nagel2021white, esser2019learned}. PTQ takes a trained high-precision NN and directly converts it to lower precision, which is easy to use and can be applied to any architecture without accessing the training pipeline or data. PTQ mainly focuses on two questions to trade off accuracy for speed: 1) Which tensor object should be quantized, e.g., NN weights or activations? 2) What is a proper quantization bit $q$ and range $[q_{min}, q_{max}]$? The concern with PTQ is that when precision $q$ is low, like 4 bits, accuracy can be low regardless of which tensor or range is chosen. QAT, on the other hand, models quantization error during training. For example, train a model with low-precision along with the original high-precision model and minimize the error between them. QAT can thus find better quantized models than PTQ in general. The cost of QAT is longer training time and data requirements.

Existing quantization methods are mostly for convolutional nets (CNNs). Quantization for GNNs, however, has its own challenges and requires special treatment. Two popular GNN quantitation methods are SGQuant and Degree-Quant, which are discussed in Appendix~\ref{app:alg_inference}. Overall, quantization is one of the easiest GNN acceleration methods to apply but shows significant speed improvement. It can also be combined with other acceleration methods to further boost speed gain. For example, \cite{zhao2020learned} combines quantization with network architecture search (NAS), where a quantization search space is specified and the best quantized GNN is selected via NAS.

\subsection{Distillation}
Knowledge distillation (KD) or simply distillation is a technique to compress ML models to smaller and thus faster models \cite{bucilua2006model}. It was shown to produce much faster models with very little, if any, accuracy loss for image classification and speech recognition \cite{hinton2015distilling}. The idea of KD is to train two models in a teacher-student fashion. The teacher outputs a probability vector for each input data with probabilities for the input to be classified into $C$ classes. The probability vectors (or a scaled version) are in $[0, 1]^C$ and are called soft labels, in contrast to the ground-truth hard labels in $\{0, 1\}^C$. Such soft labels are used to train a simpler student model. Specifically, the standard KD loss on a data instance $i$ is computed as the distance between the teacher output (soft label), i.e., $\hat \vy_i^{t}$, and the student output, i.e., $\hat \vy_i^{s}$. Two outputs are matched by minimizing the distance to transfer knowledge from the teacher to the student. Since both outputs are probability vectors, a common choice of the distance metric is the Kullback-Leibler (KL) divergence denoted as $D_{KL}(\hat \vy_i^{t} || \hat \vy_i^{s}) = \sum_{j} \hat \vy_{ij}^{t} \log \frac{\hat \vy_{ij}^{t}}{\hat \vy_{ij}^{s}}$.

In most cases, the teacher can be trained alone first, and the student is trained next fixing the teacher. KD can be naturally applied in a semi-supervised setting, a common setting for graph node classification, by training the teacher with labeled data and generating soft labels for unlabeled data to train the student~\cite{noisy_student}. KD has been widely used for compressing and accelerating DNNs like CNNs~\cite{hinton2015distilling}, and recently it has been extended to GNNs as well \cite{yang2020distilling, tinygnn_kdd20, zhang2022graphless}. The GNN KD methods discussed below mostly follow the KD framework above and mainly differ in two perspectives: 1) What are the teacher and student models? 2) What is the KD objective?

\textbf{LSP} \cite{yang2020distilling} explores KD for accelerating GNNs. Targeting node classification, LSP considers a GAT as the teacher and a GAT with fewer parameters as the student. LSP proposes a new KD objective termed local structure preserving (LSP) loss. The intuition is that node representations are generated by aggregating messages within the local structure, so the desired distillation function should transfer not only the aggregation result but also the local structure knowledge. LSP defines the local structure as the similarity between each node to its neighbors and encourages the local structure of the student to match the teacher. Specifically, the local structure of node $i$ is represented as $\vs^{(i)} \in \R^{d(v)}$, with $\vs^{(i)}_{j}$ being the local influence of node $j$ to node $i$ as $\vs^{(i)}_{j} = \exp(||\vh_i - \vh_j||_2^2) / \sum_{k: (k, i) \in \E} \exp(||\vh_i - \vh_k||_2^2)$. Plug-in node representations $\vh$ of the teacher (student) gives the teacher (student) local structure $\vs^{(i), t}$ ($\vs^{(i), s}$). Then, the local structure preserving loss $\mathcal{L}_{i}^{LSP}$ is defined using the KL divergence, similarly as the $\mathcal{L}_{i}^{KD}$. The difference being $\mathcal{L}_{i}^{LSP}$ computes the KL-divergence of $\vs^{(i), s}$ against $\vs^{(i), t}$,
\begin{equation} \label{eq:lsp_loss}
    \mathcal{L}_{i}^{LSP} = D_{KL}(\vs^{(i), s} || \vs^{(i), t}) = \sum_{j \in \N(i)} \vs^{(i), s}_{j} \log \frac{\vs^{(i), s}_{j}}{\vs^{(i), t}_{j}}.
\end{equation}

\textbf{GLNN} \cite{zhang2022graphless} explores KD for accelerating GNNs by making the student model as simple as a ``graph-less neural network'', i.e., an MLP. GLNN follows the KL divergence loss $\mathcal{L}_{i}^{KD}$, but using an MLP student model (a zero-layer GNN) with no node dependency. GLNN shows that even with a student as simple as an MLP, KD can help the student MLP achieve competitive performance to the teacher GNN on many benchmark datasets, even when used for inductive prediction on unseen new nodes. GLNN can accelerate GNNs to orders of magnitude faster because node dependency is dropped for the student. It is also shown to work with different teacher GNNs including GCN, GAT, GraphSAGE, etc. However, GLNN is not a universal solution. The analysis in \cite{zhang2022graphless} shows that GLNN fails when the conditional mutual information between node features $\mX$ and node labels $\mY$ given graph structure $\mA$ is low. Intuitively, if node labels depend mainly on graph structure but not node features, then MLPs won't be able to classify nodes even with KD.

Overall, the methods above apply KD to GNNs by adapting graph properties from different perspectives, including new objectives and new student designs. They show exciting improvements over the teacher GNN, but this line of research is still at an early stage. These KD methods were developed under different settings. Thus, we are unaware of the applicability of these distillation techniques to each other's setting and whether we can combine them. Since KD in general is a trade-off between model accuracy and speed, which method or combination will give the best result balancing these two perspectives is unknown. Also, under what exact conditions simple methods like GLNN will be applicable is unknown. These open questions make KD a promising direction to explore, e.g.,\cite{guo2023linkless, qin2025weak} study these questions for link prediction. More works are applying KD on graphs~\cite{tinygnn_kdd20, cpf, chen2021self, zheng2021cold}. Different from the methods introduced above, these works often choose students that are not necessarily simpler than the teacher, which can sometimes outperform the teacher in accuracy, but be even slower than the teacher. We discuss them in Appendix~\ref{app:alg_inference}.

\subsection{Combine Together}
Inference acceleration techniques above have been combined together for better results. For example, quantization is combined with distillation in \cite{bahri2021binary}, where GNN weights and activations are excessively quantized to binary. This is very ambitious and hard to achieve in one-step. \cite{bahri2021binary} uses a cascaded distillation scheme to gradually quantize more parts of the GNN and distill knowledge step-by-step to the next more quantized GNN down the sequence. The final quantized-and-distilled binary GNN can achieve 2X speedup on Raspberry Pi 4B with only moderate accuracy loss.

\section{GNN Acceleration: COTS Systems}\label{sec:method_system}
Beyond efficient algorithms for GNN training/inference, optimizing the system is essential for improving end-to-end throughput. Existing works accelerate GNN systems in three ways: GPU kernel acceleration, user-defined function (UDF) optimization, and scalable training systems. Unlike most DNNs, GNN computation has a message-passing paradigm requiring efficient sparse operations, which existing DNN platforms cannot sufficiently handle.

To better understand the landscape of COTS-based acceleration, we organize these techniques into a hierarchy based on the scope of optimization and the bottleneck addressed. 
Kernel-Level Optimization is the lowest level that focuses on the atomic units of computation, primarily Sparse Matrix-Matrix Multiplication (SpMM). These optimizations target compute-bound and latency-bound scenarios where the irregularity of graph data causes poor thread utilization and memory access patterns on general-purpose hardware.
UDF-Level Optimization moves up the stack and targets the flexibility bottleneck. Since GNN research rapidly produces new aggregation logic, hand-writing kernels for every variant is infeasible. These techniques optimize the execution plan for custom logic defined in high-level languages, balancing programmability with efficiency.
System-Level Optimization is the highest level that addresses memory and I/O bottlenecks inherent in processing large graphs that exceed single-device memory. These systems manage data movement, partitioning, and communication protocols across distributed hierarchies, ensuring vertical (data size) and horizontal (device count) scalability. Currently, two widely used frameworks for GNNs are PyG~\cite{pyg} and DGL~\cite{wang2019dgl}.
The following subsections discuss these three levels in detail. Table 2 summarizes kernel acceleration methods, while Tables 3 and 4 categorize system-level approaches.

\subsection{GPU Kernel Acceleration}
GPUs are powerful accelerators for DNN training and inference. GPU kernels are specialized programs optimized for parallel execution on GPU processing units, enabling faster computations. However, accelerating GNNs with GPUs remains challenging due to unique sparsity and irregularity in graphs. Table~\ref{tab:sys_kernel} summarizes recent works targeting GNN kernels, with details in Appendix~\ref{app:system}.

\begin{table*}[t]
\centering
\scriptsize
\caption{Summary of GPU Kernel Acceleration Methods.}
\resizebox{\textwidth}{!}{%
\begin{tabular}{@{}l c c | l c c@{}}
\toprule
\multirow{1}{*}{\textbf{Work}} & \multirow{1}{*}{\textbf{Methods}} & \multirow{1}{*}{\textbf{Benefits}} &
\multirow{1}{*}{\textbf{Work}} & \multirow{1}{*}{\textbf{Methods}} & \multirow{1}{*}{\textbf{Benefits}} \\ \midrule
PCGCN \citep{tian2020pcgcn} & \makecell{Workload Reorder, \\ Dual-Mode Computation} & \makecell{Better Data Locality, \\Coalesced Memory Access} &
FusedMM \citep{rahman2021fusedmm} & \makecell{Kernel Fusion} & \makecell{Balanced Workload, \\ Saved Memory} \\
\hline
fuseGNN \citep{chen2020fusegnn} & \makecell{Kernel Fusion, \\ Workload Reorder, \\ Dual-Mode Computation} & \makecell{Decreased Memory Traffic, \\ Saved Memory} &
Zhang et al.~\cite{zhang2022understanding} & \makecell{Workload Reorder, \\ Kernel Fusion, \\ Recomputing Embedding} & \makecell{Reduced Computation, \\ Reduced I/O, \\ Saved Memory} \\
\hline
TLPGNN \citep{fu2022tlpgnn} & \makecell{Workload Reorder, \\ Kernel Fusion, \\ Dynamic Workload Assignment} & \makecell{Decreased Memory Traffic, \\ Coalesced Memory Access, \\ Balanced Workload} &
MaxK-GNN \citep{peng2024maxk} & \makecell{Custom Sparse Format, \\ Workload Reorder, \\ Kernel Fusion} & \makecell{Decreased Memory Traffic, \\ Balanced Workload, \\ Reduced I/O} \\
\bottomrule
\end{tabular}
}
\label{tab:sys_kernel}
\vskip -0.1in
\end{table*}

Overall, workload reordering is necessary for identifying GPU-friendly processing patterns. Most works sequentially compute output nodes and process features in parallel. The former enables kernel fusion and avoids branch divergence while the latter guarantees coalesced memory accesses. Because graph structure can be irregular and in different formats, some works leverage dual-mode computation and dynamic workload assignment for handling different graphs.

\subsection{User-defined Function Optimization}

Flexibility in selecting message, aggregation, and update functions has been critical to GNN success under the message-passing paradigm. GNN systems enable users to natively express these functions using tensor operators in Python and execute them through system APIs. This programming paradigm is termed UDFs. Once defined, they can be reused throughout the program, increasing modularity and reducing duplication. Recently, several works ease deployment of user-defined GNNs: FeatGraph~\citep{hu2020featgraph}, Seastar~\citep{wu2021seastar}, Graphiler~\citep{xie2022graphiler}. Specifically, gSampler~\citep{gong2023gsampler} develops a GPU-based programming model for defining new sampling methods. Details are in Appendix~\ref{app:system}.

Existing automation tools for GNN computation are limited because concurrent deep learning (DL) backends do not well support sparse operations. This direction is maturing as several recent backends for sparse DL operations have been proposed. For example, SparseTIR~\citep{ye2022sparsetir} is a general compiler infrastructure for sparse DL operations, developing intermediate representations that can represent operators supported by Seastar, FeatGraph, and Graphiler.

\subsection{Scalable Training Systems}

To train GNNs on large graphs, two types of training acceleration systems have been developed:

\begin{itemize}
    \item \textbf{Vertical scalability} handles \textit{scaling data} with restricted computation resources, which is highly practical for large training graphs. Systems pursuing vertical scalability typically assume low-cost or unlimited host memory.
    \item \textbf{Horizontal scalability} expects better efficiency with \textit{scaling resources}. However, this is challenging because more workers incur more cross-device dependency, leading to greater communication and synchronization overhead that degrades training throughput.
\end{itemize}

\begin{table}[t]
\center
\small
\caption{Comparison of the Three Types of Scalable GNN Training Systems.}
\begin{tabular}{ccc}
\toprule
\textbf{System Type} & \textbf{Main Focus} & \textbf{Challenges and Contribution} \\
\midrule
On-device system & Full-graph aggregation & Communication and memory pattern \\
\hline
Swap-based system & Full-graph aggregation & Workload scheduler \\
\hline
Sampling-based system & Sampling-based aggregation & Data movement and caching strategy \\
\bottomrule
\end{tabular}
\label{tab:sys_scope}
\vskip -0.1in
\end{table}

\begin{figure}[t]
\begin{center}
\includegraphics[width=0.8\textwidth]{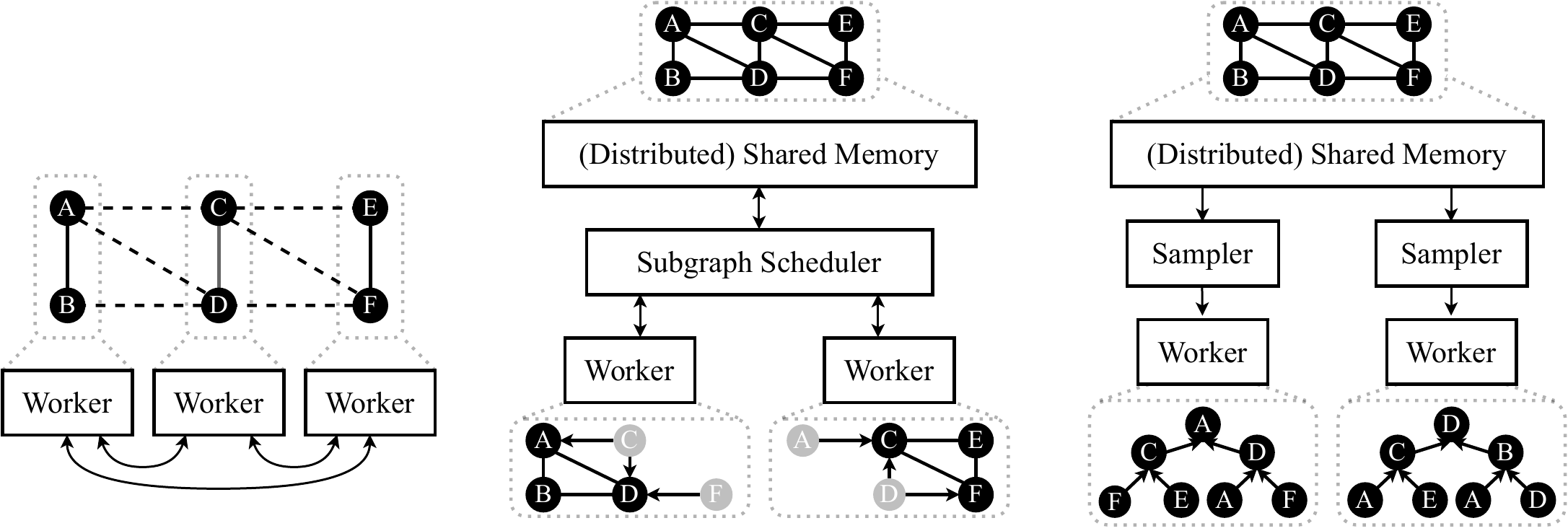}
\end{center}
\captionsetup[subfloat]{skip=-1em}
    \subfloat[\label{fig:scalable-a}On-device Systems]{\hspace{.3\linewidth}}
    \subfloat[\label{fig:scalable--b}Swap-based Systems]{\hspace{.3\linewidth}}
    \subfloat[\label{fig:scalable-c}Sampling-based Systems]{\hspace{.3\linewidth}}
\caption{Overview of three types of scalable GNN training systems. Dashed lines in (a) indicate dependency between nodes from different workers, while grey nodes in (b) serve only as input nodes for subgraph computations. Model synchronization among workers is not shown for visual clarity.}
\Description{The figure shows a three-layer distributed system. Top and bottom layers consist of Distributed Shared Memory storing graph partitions A through F. The middle layer contains a Subgraph Scheduler, multiple Samplers, and a cluster of Workers. Data flows from the shared memory through the samplers, which extract subgraphs, and then to the workers for parallel GNN training.}
\vskip -0.15in
\label{figure:scalable_framework}
\end{figure}

For recent distributed GNN training systems, \citet{shao2022distributed} classifies works by hardware facilities (e.g., multi-GPU, GPU cluster, CPU cluster), while \citet{lin2022comprehensive} categorizes them as full-batch or mini-batch training. This work focuses on scalable GNN training, a wider topic than distributed training, including efficient distributed training (horizontal scalability) and large-scale training with limited resources (vertical scalability). We observe that key contributions of each system overlap among categories in existing surveys, hindering understanding of technical differences. To facilitate clearer comparison, we categorize systems based on how data is loaded onto computation devices: on-device systems, swap-based systems, and sampling-based systems, with demonstrations in Figure~\ref{figure:scalable_framework} and comparison in Table~\ref{tab:sys_scope}. 

\subsubsection{On-device Systems} 
(Figure~\ref{figure:scalable_framework} (a)) distributedly store the entire graph and features on computation devices (e.g., GPUs or CPUs). Since each worker maintains only a portion of data, these systems focus on full-graph aggregation for better utilizing computation resources. During training, each worker transfers intermediate embeddings to support full-graph aggregation in the next layer.
This design resembles distributed data parallelism in DNN training but has two fundamental differences: 1) GNN requires feature transfer for each layer, exploding communication volume; and 2) node features/embeddings are substantially larger than the model. These unique communication and memory challenges motivate various training systems, also termed \textit{partition parallelism}.

\textbf{PipeGCN}~\cite{wan2022pipegcn} is one example on-device system. It optimizes straightforward partition parallelism through pipelining embedding computation and dependent communication across training iterations. For distributed full-batch training, each worker typically waits for dependent neighbors' embeddings or gradients before performing local forward/backward computation. In PipeGCN, workers use transferred embeddings/gradients from the previous iteration. This breaks the computation-communication dependency and allows parallel computation and communication. Although model updates deviate from traditional gradient descent due to stale embeddings and gradients, convergence analysis shows PipeGCN achieves $\mathcal{O}(T^{-\frac{2}{3}})$ convergence rate, with $T$ being training iterations, better than sampling-based methods ($\mathcal{O}(T^{-\frac{1}{2}})$). PipeGCN further employs embedding/gradient momentum to mitigate errors from stale data transfer. Other on-device systems include CAGNET~\cite{tripathy2020reducing}, Dorylus~\cite{thorpe2021dorylus}, $\mathbf{P^3}$~\cite{gandhi2021p3}, LLCG~\cite{ramezani2022learn}, BNS-GCN~\cite{wan2022bns}, SAR~\cite{mostafa2022sequential}, and Sancus~\cite{peng2022sancus}. Table~\ref{tab:sys_works} summarizes their key properties with detailed discussion in Appendix~\ref{app:system}.

\subsubsection{Swap-based Systems.}
(Figure~\ref{figure:scalable_framework} (b)) store all data in (distributed) shared memory and swap it and intermediate results to workers for computation. This enables each node to fully utilize neighbors' embeddings. However, developing a workload scheduler ensuring vertical scalability poses two challenges: 1) balancing workload among GPUs is non-trivial, and 2) naive full-graph aggregation with limited devices incurs long iteration latency as it only updates the model once per epoch. Effective solutions are crucial for improving efficiency.

\textbf{NeuGraph}~\citep{ma2019neugraph} and \textbf{Roc}~\citep{jia2020improving} are pioneering works for GNN training with scaling graphs. Given restricted GPU resources, they process output nodes sequentially and store all intermediate results in host CPU DRAM for backward propagation computation. Both support multi-GPU GNN training. NeuGraph only supports distributed training in a single machine, adopting equal-vertex partition for heuristic workload balancing. Roc strengthens scalability by supporting multi-machine training and leverages a learnable cost model for partitioning graphs toward better workload balance. Other popular systems include GNNAdvisor~\citep{wang2021gnnadvisor}, GNNAutoScale~\citep{fey2021gnnautoscale}, and GraphFM~\citep{yu2022graphfm}. We summarize their key properties in Table~\ref{tab:sys_works} with detailed discussion in Appendix~\ref{app:system}.

\subsubsection{Sampling-based Systems} 
(Figure~\ref{figure:scalable_framework} (c)) save the whole graph and features in (distributed) shared memory but adopt neighbor sampling instead of full-neighbor aggregation. Each sampler first determines sampled input nodes based on output nodes for mini-batch training, then fetches features from (distributed) shared memory. The primary challenge is data movement for batch preparation. These systems focus on developing better data movement and caching strategies to optimize communication. Overall, sampling-based systems enhance both horizontal and vertical scalability, making them promising for GNN training on large graphs.

\textbf{AliGraph} (also known as Graph-Learn)~\cite{zhu2019aligraph} is a CPU-based distributed training framework for a wide range of GNN variants and training algorithms. For efficient large-scale graph training, AliGraph utilizes distributed storage with cached important nodes and an optimized data sampler for fetching dependent node features. The system implements four graph partition algorithms for different tasks (e.g., METIS for sparse graphs, streaming-style partition for dynamic graphs with frequent edge changes). Since distributed storage increases data access overhead, AliGraph caches frequently visited nodes to avoid repetitive data transfer. The system also implements three sampling strategies to speed up training and accelerates samplers with dynamic sampling weights. 
Other popular sampling-based systems include PaGraph~\cite{lin2020pagraph}, 
DistDGL~\cite{zheng2020distdgl} and DistDGLv2~\cite{zheng2021distributed}, 
SALIENT~\cite{kaler2022accelerating}, GNNLab~\cite{yang2022gnnlab}, BGL~\cite{liu2021bgl}, GNS~\cite{dong2021global}, MariusGNN~\citep{waleffe2022marius++}, and DSP~\cite{cai2023dsp}.
We summarize their key properties in Table~\ref{tab:sys_works} with detailed discussion in Appendix~\ref{app:system}. 

Overall, the core challenge toward scalable GNN training systems is data transfer. To address this, several methods have been widely adopted: caching frequently visited nodes to avoid redundant communication, using asynchronous communication to hide overhead, and leveraging stale embeddings to reduce communication volume. However, saving memory and balancing workload are also critical factors when designing GNN training systems. Achieving these goals requires combining effective algorithms and system optimizations for efficient training on large graphs.

\begin{table}
\centering
\scriptsize
\caption{Summary of Scalable Training Systems.}
\begin{tabular}{cccccc}
\toprule
\thead{\textbf{System}\\\textbf{Type}} & \textbf{System} & \thead{\textbf{Computation}\\\textbf{ Hardware}} & \textbf{Aggregation} & \thead{\textbf{Adjust}\\\textbf{ Algorithm}}  & \thead{\textbf{Open} \\ \textbf{ Source}} \\ 
\midrule
\multirow{8}{*}{\thead{On-device\\ System}} 
& CAGNET~\cite{tripathy2020reducing}         & GPU & Full-graph           & \xmark & \checkmark \\
& Dorylus~\cite{thorpe2021dorylus}        & Hybrid & Full-graph         & \xmark & \checkmark \\
& $P^3$~\cite{gandhi2021p3}          & GPU & Sampling             & \checkmark & \xmark \\
& PipeGCN~\cite{wan2022pipegcn}        & GPU & Full-graph           & \checkmark & \checkmark \\
& LLCG~\cite{ramezani2022learn}           & GPU & Full-graph           & \checkmark & \checkmark \\
& BNS-GCN~\cite{wan2022bns}        & GPU & Hybrid             & \checkmark & \checkmark \\
& SAR~\cite{mostafa2022sequential}            & CPU & Full-graph           & \xmark & \checkmark \\
& Sancus~\cite{peng2022sancus}         & GPU & Full-graph           & \checkmark & \checkmark \\ 
\midrule

\multirow{6}{*}{\thead{Swap-based\\ System}} 
& NeuGraph~\cite{ma2019neugraph}       & GPU & Full-graph           & \xmark & \xmark \\
& Roc~\cite{jia2020improving}            & GPU & Full-graph           & \xmark & \checkmark \\
& GNNAdvisor~\cite{wang2021gnnadvisor}     & GPU & Full-graph           & \xmark & \checkmark \\
& GNNAutoScale~\cite{fey2021gnnautoscale}   & GPU & Full-graph           & \checkmark & \checkmark \\
& GraphFM~\cite{yu2022graphfm}       & GPU & Full-graph           & \checkmark & \checkmark \\
\midrule

\multirow{8}{*}{\thead{Sampling-based\\ System}}
& AliGraph~\cite{zhu2019aligraph}       & CPU & Sampling             & \checkmark & \checkmark \\
& PaGraph~\cite{lin2020pagraph}        & GPU & Sampling             & \xmark & \checkmark \\
& DistDGL~\cite{zheng2020distdgl}        & CPU & Sampling             & \xmark & \checkmark \\
& DistDGLv2~\cite{zheng2021distributed}      & GPU & Sampling             & \xmark & \xmark \\
& SALIENT~\cite{kaler2022accelerating}        & GPU & Sampling             & \xmark & \checkmark \\
& GNNLab~\cite{yang2022gnnlab}         & GPU & Sampling             & \xmark & \checkmark \\
& BGL~\cite{liu2021bgl}             & GPU & Sampling             & \xmark & \xmark \\
& GNS~\cite{dong2021global}            & GPU & Sampling             & \checkmark & \xmark \\     
& MariusGNN~\cite{waleffe2022marius++}      & GPU & Sampling             & \checkmark & \checkmark \\             
\bottomrule
\vspace{-0.1in}
\end{tabular}\label{tab:sys_works}
\end{table}

\subsection{Guidelines for Design of COTS-Based GNN Systems}
The scalability challenge of GNN training and inference on COTS systems can only be effectively addressed through deliberate collaboration across algorithm, system, and hardware levels. While this section surveys a wide range of individual techniques---specialized GPU kernels, UDF optimizations, and scalable distributed frameworks---these contributions are most powerful when combined in principled ways. Isolated optimizations often leave significant performance on the table by failing to account for how choices at one level create opportunities or bottlenecks at others. This subsection provides practical, actionable guidelines for designing high-performance COTS-based GNN systems, with concrete examples illustrating successful multi-level co-design.

\subsubsection*{1. Start with Workload and Bottleneck Characterization}

Before selecting any technique, quantitatively profile the target workload:

\begin{itemize}[leftmargin=2em]
    \item \textbf{Graph properties}: number of nodes/edges, degree distribution, diameter, feature dimension.
    \item \textbf{Model properties}: depth, aggregation function, training regime (full-batch vs. mini-batch).
    \item \textbf{Execution bottlenecks}: use NVIDIA Nsight, PyTorch Profiler, or DGL/PyG built-in profiling to determine whether the system is memory-bandwidth-bound, compute-bound, communication-bound, or kernel-launch-bound.
\end{itemize}

This characterization is essential because optimal design space changes dramatically across regimes. For example, power-law graphs with high diameter are memory-bound on GPUs due to random accesses, whereas dense graphs with high feature dimensions are compute-bound.
\subsubsection*{2. Choose the Algorithmic Family that Best Matches Hardware Constraints}
Algorithmic choices are the most flexible lever and should be selected \textit{early} to enable system and hardware efficiency.
\begin{itemize}[leftmargin=2em]
\item \textbf{Memory-constrained devices ($\leq$40 GB HBM, e.g., A40)}: Use mini-batch training with sampling (e.g., GraphSAGE) or cluster-based methods (e.g., ClusterGCN), which dramatically reduce working set size and enable high GPU utilization even on consumer hardware.
\item \textbf{High-memory devices ($\geq$80 GB, e.g., H100)}: Prefer full-graph or large-subgraph methods when the graph fits, or swap-based methods (e.g., NeuGraph) when it marginally exceeds memory. Exact aggregation yields faster convergence and eliminates sampling variance.
\item \textbf{Large graphs exceeding single device}: Favor sampling distributed training (e.g., DistDGL) or partitioned full-graph methods with feature momentum or staleness tolerance (e.g., Roc).
\end{itemize}

\subsubsection*{3. Apply System-Level Optimizations that Exploit the Chosen Algorithm}
Once the algorithmic family is fixed, select system techniques known to synergize with it:
\begin{itemize}[leftmargin=2em]
\item \textbf{Always enable maximum kernel fusion}: Unified thread mapping (e.g., fuseGNN) reduces global memory traffic by 2-5$\times$. Fusion is especially critical with high-dimensional features.
\item \textbf{Exploit caching in sampling-based training}: Caching high-degree or historically frequent nodes in GPU memory (e.g., PaGraph) to increase effective batch size by 4-8$\times$.
\item \textbf{Minimize communication volume and overlap with computation}: Use METIS or learnable partitioning (Roc) to reduce edge-cut, then apply pipelining (e.g., PipeGCN) or asynchronous execution to hide latency on NVLink/InfiniBand interconnects.
\item \textbf{Use sparse-aware compilers/backends when possible}: SparseTIR, or fused SpMM/SDDMM kernels, provide automatic fusion and layout optimization.
\end{itemize}

\subsubsection*{4. Decision Framework for System Selection/Design}

\begin{table}[tb]
\footnotesize
\centering
\caption{Recommended COTS-based GNN system design choices by regime.}
\label{tab:guidelines}
\resizebox{\textwidth}{!}{
\begin{tabular}{llll}
\toprule
\textbf{Regime} & \textbf{Graph size} & \textbf{Preferred algorithm} & \textbf{Key system techniques} \\
\midrule
Single consumer GPU & Any & Neighbor/cluster sampling & \makecell[l]{Fusion + vertex caching} \\
\midrule
Single high-end GPU & 200M--1B edges & Full-graph or large-subgraph & \makecell[l]{Max fusion, SpMM/GeMM auto-switch} \\
\midrule
Multi-GPU (4--16) & 1B--10B edges & Sampling + pipelining & \makecell[l]{Cache + async + pipeline} \\
\midrule
CPU$+$GPU or large cluster & $>$10B edges & Partitioned & \makecell[l]{Full-graph or advanced sampling, \\ Feature momentum, staleness} \\
\bottomrule
\end{tabular}
}
\vspace{-0.1in}
\end{table}

Table~\ref{tab:guidelines} summarizes our recommendations. The key principle is to \textit{push as much exact computation as memory and bandwidth allow}, then use approximation and caching to handle the remainder.
\subsubsection*{5. Key Trade-offs and Practical Advice}

\begin{itemize}[leftmargin=2em]
    \item \textbf{Accuracy vs. throughput}: Sampling and staleness introduce bias/variance; budget extra epochs or use variance-reduction/control-variate methods.
    \item \textbf{Implementation complexity}: Start with PyG/DGL + FusedMM/fuseGNN (easy, 3--8$\times$ speedup), then add caching/pipelining only if needed.
    \item \textbf{Cost-effectiveness}: For graphs $>$10B edges, CPU-based systems (Dorylus) or disk-backed systems (MariusGNN) are often cheaper than GPU clusters despite lower per-epoch speed.
\end{itemize}

By systematically applying these guidelines---characterizing the workload, selecting hardware-aware algorithms first, then layering synergistic system optimizations---practitioners can achieve end-to-end speedups over naïve implementations while maintaining convergence close to exact training. Future COTS-based GNN systems should automate these co-design decisions through learned schedulers and compilers, further closing the gap to custom hardware solutions.

\section{GNN Acceleration: Customized Hardware}\label{sec:method_hardware}
The increasing interest in GNNs has led to the development of customized accelerators (FPGA or ASIC) for faster processing. While GNNs share similarities with CNNs in network architecture, differences in computation complexity and communication patterns make numerous CNN accelerators~\cite{flexcnn, zhang2018dnnbuilder, eyeriss} unsuitable for GNNs. Specifically, GNNs require Matrix Multiplication (MM) units and have irregular memory access due to the unstructured nature of graphs. Although both the Aggregation and Update stages can be modeled as MMs, their computation and communication patterns differ. The Aggregation stage handles an ultra-sparse adjacency matrix, while the Update stage matrices are dense or have much lower sparsity. Consequently, using only dense MM units~\cite{autosa} or sparse MM (SpMM) units~\cite{srivastava2020matraptor, song2022sextans} is inefficient. Furthermore, GNNs work with vectors assigned to each node, unlike traditional graph algorithms that use scalar values, resulting in distinct computation and communication requirements. These distinctions have motivated researchers to design specialized hardware modules for GNNs.

Abadal et al.~\cite{abadal2021computing} concisely review the endeavors in this domain, offering insightful perspectives. 
We accentuate the challenges and differences in employed optimization techniques more clearly. Section~\ref{subsec:hw_challenges} describes the challenges designers face and decisions required in developing accelerators. Section~\ref{sec:customized_hw} reviews the most prominent accelerators proposed to date.

\subsection{Challenges in Customized Hardware Design} \label{subsec:hw_challenges}
Graphs and GNNs possess specific characteristics requiring special attention in customized hardware design. Designing accelerators for them requires consideration of unique features:

\textbf{Wide range of GNN layers.}
Various GNN models utilize distinct Aggregation and Update methods, potentially incorporating graph/subgraph pooling layers and non-linear activation functions. These differences may impact execution bottlenecks and areas requiring acceleration, resulting in a trade-off between flexibility and performance.
Supporting multiple computation modes necessitates generality, which may prevent optimal performance for each specific case.

\textbf{Computation/communication pattern and sparsity rate disparity.}
Computation and communication requirements of different GNN steps vary significantly due to distinct sparsity rates and irregular access patterns. For instance, the adjacency matrix is ultra-sparse, whereas the weight matrix is dense. Node embeddings are generally dense but may become sparse when generated by activation functions like ReLU, though with much lower sparsity than the adjacency matrix. Additionally, end-to-end GNN applications may utilize other computation patterns such as MLPs for processing node/graph embeddings. These differences pose significant challenges to hardware designers, who must address the following sources of disparity:
\begin{itemize}[leftmargin=2em]
	\item Graph (adjacency matrix) sparsity: Graphs are often large, ultra-sparse matrices, posing memory design challenges. For instance, Orkut~\cite{orkut} has over 3 million nodes and 223 million edges, yet has density of $2.5\times10^{-5}$. The large size requires an on-chip and off-chip memory hierarchy, while sparsity translates to irregular accesses spanning multiple memory hierarchies. 

	\item Sparsity introduced by activation operators: Activation operators like ReLU can produce sparsity in node embeddings. To enable efficient computation, it is desirable to address both graph and feature sparsity through zero elimination or bypassing techniques in computation units, accelerating computation and avoiding wasted cycles on zero-valued operands.

	\item Coordination between modules: The differences in data layout caused by 
    sparsity necessitate the coordination of computation modules. Specifically, the 
    Aggregation step in GNNs often requires sparse matrix-vector multiplication 
    (SpMV) or SPMM modules, which typically shuffle the data layout to enable 
    efficient processing.  
The Update (also called transformation) step typically works on continuous matrix/vector layouts. Consequently, accelerators must coordinate data layout between modules. In some GNNs, weights or embeddings in the Update stage may also be sparse, requiring further coordination of computation patterns and sparsity of the 
Update module with the SpMV and SpMM modules.
\end{itemize}	

\textbf{Dynamic input and network structure.}
Given the dynamic nature of graphs, the architectural decisions for GNNs are highly dependent on the specific characteristics of the input graph and the model's hyperparameters. The memory and computation requirements of a GNN accelerator are significantly impacted by factors such as the size of the graph, its level of sparsity, and the dimensions of the vectors involved. Therefore, the features and properties of each input graph can be highly distinct from one another, including the following:
\begin{itemize}[leftmargin=2em]
    \item Different scales and structure of graphs: Real-world graphs range from a few nodes to millions. This variation affects computation resources needed for GNN computations. Studies show graph size significantly impacts scheduling and optimization technique selection \cite{sohrabizadeh2022streamgcn, geng2020awb}. Additionally, varying neighbor counts introduce workload imbalance.

    \item Vectors with various sizes per node: Traditional graph algorithms like BFS, SSSP, and PageRank assign scalar values to nodes. GNNs utilize long feature vectors, affecting memory access patterns and parallelism. Long vectors enable intra-node parallelism and data reuse when applying the same weight matrix to all node embeddings. Multiple GNN layers can operate on different vector sizes, leading to changes in access patterns and required parallelism across layers, creating customization opportunities in hardware design of GNN accelerators.
\end{itemize}

Given these special features, designers must consider how to address these critical factors when designing customized hardware accelerators: selecting appropriate data structures and architectures for large, irregular data structures; optimizing parallelism and exploiting sparsity; balancing workload across processing units; minimizing communication overheads between processing units; and adjusting data flow to avoid bottlenecks. Additionally, designers must optimize for specific GNNs and their computation/communication patterns. Given the wide range of GNN models, trade-offs between generality (flexibility), scalability, and specialization also need to be considered.

\subsection{Summary of The Existing Customized Accelerators} \label{sec:customized_hw}

The accelerators in this section employ distinct approaches to address the aforementioned challenges, influenced by the GNNs and datasets they target. Such trade-offs affect attainable peak performance and generalizability. As the application scope narrows, more customization opportunities arise, improving performance at the expense of flexibility. These accelerators primarily target inference rather than training. Constantly changing network architectures during training present challenges for hardware adaptation, so customized accelerators for training are rare~\cite{zeng2020graphact, chen2021rubik}. 

To provide a systematic view, we classify existing accelerators along two primary conceptual axes. These distinctions highlight the fundamental design trade-offs between flexibility and resource utilization.
The first major axis is the \textbf{pipeline architecture} that distinguishes accelerators based on how they map the computation flow. Unified architectures implement all stages of the GNN pipeline (e.g., aggregation, update, and readout) in a holistic manner, typically utilizing a shared computation engine for different phases. This allows for high resource reuse but requires sophisticated control logic to handle diverse compute patterns. In contrast, stage-specialized architectures employ distinct hardware modules dedicated to specific stages (e.g., a dedicated sparse engine for aggregation and a dense engine for update), enabling inter-stage pipelining and tailored optimizations for the distinct memory access patterns.
The second major axis is the \textbf{workload generality} that distinguishes accelerators based on the scope of the target utility. General-purpose platforms are designed to support a wide range of GNNs and varying aggregation functions. They prioritize programmability to accommodate the rapid evolution of GNN models. Tailored accelerators are for specific application domains or graph structures, such as specific physics simulations. By sacrificing generality, these designs can strip away control overhead and achieve higher peak efficiency.

Table~\ref{tbl:hw_comparison} summarizes current customized accelerators, comparing them based on five main features: graph size, target GNNs, support for layer customization based on dynamic input structure, approach to handle inherent sparsity, and parallelization levels employed. 
Prevalent parallelization schemes include parallelizing the feature dimension for all stages, parallelizing nodes in the Update stage, and pipelining main stages of each layer for intra-layer parallelization. Some approaches exploit parallel processing of multiple edges in Aggregation, pipeline multiple layers (inter-layer), or process multiple graphs in parallel (batch).
In the remainder of this section, related works are categorized by workload flexibility. Section~\ref{sec:customized_hw_general} summarizes notable works proposing accelerators for multiple GNNs. Another category focuses mainly on GCN operations, further customizing microarchitecture for GCNs, detailed in Section~\ref{sec:customized_hw_special}.

\begin{table*}[t]
\scriptsize
\centering
\caption{Customized Accelerators Properties}
\label{tbl:hw_comparison}
\begin{tabular}{cccccc}
\toprule
\multirow{2}{*}{\textbf{Work}} & \multirow{2}{*}{\makecell{\textbf{Graph} \\ \textbf{Size}}} & \multirow{2}{*}{\makecell{\textbf{Target} \\ \textbf{GNNs}}}  & \multirow{2}{*}{\makecell{\textbf{Layer}\\ \textbf{Customization}}} & \multirow{2}{*}{\makecell{\textbf{Sparsity} \\ \textbf{Support}}}   & \multirow{2}{*}{\textbf{Parallelization}}\\\\ \hline

\multirow{4}{*}{\makecell{EnGN \\ \cite{liang2020engn}}}
& Large, & GCN, GRN & & Reorganize edges & Intra-layer, Node-level \\
& Ultra & Gated GCN & \xmark & for aggregation & Feature-level, Edge-level \\
& Large & R-GCN \\
& & GraphSage  \\
\\
\multirow{2}{*}{\makecell{Rubik \\~\cite{chen2021rubik}}}
& Small, & GIN& \xmark & Pre-process to  & Intra-layer, Node-level \\
& Large & GraphSage && reorder the graph & Feature-level, Edge-level
\\
\\
\multirow{4}{*}{\makecell{HyGCN \\ \cite{yan2020hygcn}}}
&& GCN & &Window sliding and & Intra-layer, Node-level \\
& Large &  GraphSage &\xmark & shrinkage + sampler as & Feature-level, Edge-level  \\
&& GIN && a \textit{sparsity eliminator} & (if feature dimension is \\
&& DiffPool && tool for aggregation & small)\\
\\
\multirow{3}{*}{\makecell{FlowGNN \\ \cite{sarkar2022flowgnn}}}
& Small & GCN, GIN &  &\textit{On-the-fly multicasting} & Intra-layer, Node-level\\
&Large & GAT, PNA & \xmark & for data distribution & Feature-level, Edge-level \\
&& DGN, VN &&  &  \\
\\
\multirow{3}{*}{\makecell{BlockGNN \\ ~\cite{zhou2021blockgnn}}}
& Large & GS-Pool & & Normal SIMD process & Intra-layer, Node-level \\
& & GCN, GAT & \xmark && Feature-level, Edge-level \\
& & GGCN \\
\\
\multirow{3}{*}{\makecell{DeepBurning \\ -GL \cite{liang2020deepburning}}}
& Large & GCN & & Degree-aware caching & Intra-layer, Node-level \\
&& GS-Pool & \checkmark && Feature-level, Edge-level \\
&& EdgeConv \\
\\
\multirow{5}{*}{\makecell{AWB-GCN \\ \cite{geng2020awb}}}
&&&& Load-balancing  &\\
& & &   &for both stages:& Inter-layer, Intra-layer \\
&Large & GCN& \checkmark& \textit{distribution smoothing}, & Node-level, Edge-level\\
&&&& \textit{remote switching}, \\
&&&& \textit{evil row remapping}\\
\\
\multirow{4}{*}{\makecell{StreamGCN \\ \cite{sohrabizadeh2022streamgcn}}} 
& &&& Pre-process for & Inter-layer, Intra-layer \\
&Small & GCN & \checkmark &aggregation - In-situ  & Feature-level, Node-level\\
&&&& support for update to & Edge-level, Batch\\
&&&& prune zeros on-the-fly\\
\\
\multirow{3}{*}{\makecell{GraphACT \\ \cite{zeng2020graphact}}}
&&&& \textit{Redundancy reduction} & Intra-layer, Node-level \\
& Small & GCN &\xmark &to reduce the number of & Feature-level\\
&&&& edges for aggregation\\
\\
\multirow{3}{*}{\makecell{Zhang et al. \\ ~\cite{zhang2020hardware}}}
& && & Pre-process to partition & Intra-layer,  Node-level \\
&Large & GCN & \xmark& and reorder the graph,  & Feature-level, Edge-level\\
&&&&  and redundancy reduction \\
\\
\multirow{3}{*}{\makecell{BoostGCN \\ \cite{zhang2021boostgcn}}}
&&& & 3-D tiling of edges, & Intra-layer, Node-level\\
& Large & GCN & \xmark& nodes, and features; & Feature-level, Edge-level \\
&&&& \textit{sort-and-combine} unit \\
\\
\multirow{2}{*}{\makecell{G-CoS \\ \cite{zhang2021g}}}
& Small & GCN, GraphSage & & \textit{On-the-fly multicasting} & Intra-layer, Node-level\\
& Large & GAT, GIN & \checkmark& for data distribution & Feature-level \\
\\
\multirow{2}{*}{\makecell{GCoD \\ \cite{you2022gcod}}}
&& GCN, GIN & & Workload balanced denser, & Intra-layer, Node-level\\
& Small & GAT & \xmark& light workload sparser & Feature-level, Edge-level \\
\\
\multirow{2}{*}{\makecell{I-GCN \\ \cite{geng2021gcn}}}
& Small & GCN & & On the fly reorder & Intra-layer, Node-level \\
& Large & GraphSage & \checkmark&   and islandize the graph& Feature-level \\
\bottomrule
\end{tabular}
\vskip -0.1in
\end{table*} 
\subsection{Accelerators for General Workloads} \label{sec:customized_hw_general}
A fundamental design decision is determining the range of applications to support. This section overviews works proposing flexible accelerators for multiple GNNs. These works either create a unified architecture handling all computation stages (Section~\ref{sec:customized_hw_unified}), or develop specialized engines for primary computation stages due to distinct computation/ communication patterns (Section~\ref{sec:customized_hw_dedicated}).

\subsubsection{All-Stage Unified Architecture.} \label{sec:customized_hw_unified}
These works propose a unified architecture to process multiple computation stages despite their division into distinct phases. The proposed accelerators develop a singular engine handling these stages, even with different computation and communication requirements. They are \textbf{EnGN} \cite{liang2020engn}, \textbf{Rubik}~\cite{chen2021rubik}, \textbf{G-CoS} \cite{zhang2021g}, \textbf{GCoD} \cite{you2022gcod}, and \textbf{I-GCN} \cite{geng2021gcn}. We summarize their main characteristics in Table~\ref{tab:hw_performance} and elaborate details in Appendix~\ref{app:hardware}.

\subsubsection{Dedicated Module per Stage.} \label{sec:customized_hw_dedicated}
In contrast to the previous category, these works propose specialized engines for primary computation stages. This approach allows each engine to be customized according to stage requirements and facilitates concurrent execution of different stages. Representative works include \textbf{HyGCN}~\cite{yan2020hygcn}, \textbf{FlowGNN} \cite{sarkar2022flowgnn}, \textbf{BlockGNN}~\cite{zhou2021blockgnn}, and \textbf{DeepBurning-GL}~\cite{liang2020deepburning} summarized in Table~\ref{tab:hw_performance}. Details are deferred to Appendix~\ref{app:hardware}.

\subsection{Accelerators for Special Workload} \label{sec:customized_hw_special}
This literature focuses on more specialized algorithms, concentrating on a singular GNN algorithm (mostly GCN) and tailoring solutions accordingly. Given that target models may contain multiple layers, the first subgroup develops a deep pipeline with custom layer modifications (Section~\ref{sec:customized_hw_custom_layer}). The second subgroup adopts a fixed hardware approach for all layers (Section~\ref{sec:customized_hw_fixed_layer}).

\subsubsection{Layer-Customizable Deep Pipeline.} \label{sec:customized_hw_custom_layer}
These studies argue that GCN accelerator optimization should reduce memory transaction expenses. They construct deep pipelines circumventing off-chip memory transactions for intermediate results among GCN layers. By allocating specific engines for each layer, they can tailor hardware parameters based on individual layer workloads. Two representative works are AWB-GCN \cite{geng2020awb} and StreamGCN \cite{sohrabizadeh2022streamgcn}, found in Table~\ref{tab:hw_performance} and Appendix~\ref{app:hardware}.

\subsubsection{All-Layer Unified Architecture.} \label{sec:customized_hw_fixed_layer}
A GCN has multiple layers with distinct features (e.g., node embedding dimension).
In contrast to Section~\ref{sec:customized_hw_custom_layer}, studies in this category propose constructing more adaptable architectures across layers, utilizing the same engines for all layers. 
Since these works focus on specific workloads (GCN), they offer greater customization possibilities than those in Section~\ref{sec:customized_hw_general}, as indicated in GraphACT \cite{zeng2020graphact}, Zhang et al.~\cite{zhang2020hardware}, and BoostGCN~\cite{zhang2021boostgcn}. We summarize them in Table~\ref{tab:hw_performance} and defer detailed discussion to Appendix~\ref{app:hardware}.

\begin{table*}[t]
\scriptsize
\centering
\caption{Performance and Validation Methodology of the Customized Accelerators}
\label{tbl:hw_perf}
\begin{tabular}{cccc}
\toprule
\multirow{1}{*}{\textbf{Work}} & Validation Methodology & \makecell{End-to-end \\ Measurement}& Reported Speedup \\ \hline
\\
\multirow{2}{*}{\makecell{EnGN \\ \cite{liang2020engn}}}
& ASIC (TSMC 14nm with HBM 2.0 at 1GHz) & \xmark & $1802.9\times$, $19.75\times$, and $2.97\times$ over CPU \\
& cycle-accurate simulator &&  (PyG $\&$ DGL), GPU (PyG $\&$ DGL), and HyGCN
\\
\\
\multirow{2}{*}{\makecell{Rubik \\~\cite{chen2021rubik}}}
& ASIC (45nm at 500MHz) & \xmark & $3.42-46.7\times$, and $1.3-14.16\times$ over\\
& cycle-accurate simulator && PyG-GPU, and Eyeriss~\cite{eyeriss}-like accelerator
\\
\\
\multirow{2}{*}{\makecell{HyGCN \\ \cite{yan2020hygcn}}}
& ASIC (TSMC 12nm with HBM at 1GHz) & \xmark & $1509\times$, and $6.5\times$ over\\
& cycle-accurate simulator && PyG-CPU, and PyG-GPU
\\
\\
\multirow{2}{*}{\makecell{FlowGNN \\ \cite{sarkar2022flowgnn}}}
& FPGA (Xilinx Alveo U50 at 300MHz) & \checkmark & $24-254\times$, $1.3-477\times$, and $1.26\times$ over\\
& on-board evaluation & & PyG-CPU, PyG-GPU, and I-GCN
\\
\\
\multirow{2}{*}{\makecell{BlockGNN \\ ~\cite{zhou2021blockgnn}}}
& FPGA (Xilinx ZC706 at 100MHz) & \xmark & $2.3\times$, and $4.2-8.3\times$ over \\
& on-board evaluation and analytical modeling & & CPU (TensorFlow-based), and HyGCN
\\
\\
\multirow{2}{*}{\makecell{DeepBurning \\ -GL \cite{liang2020deepburning}}}
& FPGA (Xilinx ZC706, KCU1500, and Alveo U50) & \xmark & $7.43-346.98\times$, $3.7-16.5\times$, and $6.28\times$ over\\
& on-board evaluation at 100, 200, and 200MHz && DGL-CPU, DGL-GPU, and HyGCN on FPGA
\\
\\
\multirow{2}{*}{\makecell{AWB-GCN \\ \cite{geng2020awb}}}
& FPGA (Intel Stratix 10 SX) & \xmark & $3255\times$, $80.3\times$, $5.1\times$ over\\
& on-board evaluation && PyG-CPU, PyG-GPU, and HyGCN
\\
\\
\multirow{2}{*}{\makecell{StreamGCN \\ \cite{sohrabizadeh2022streamgcn}}} 
& FPGA (Xilinx KU15P, Alveo U50 and U280) & \checkmark & $8.2-18.2\times$, and $12.1-26.9\times$ over\\
& on-board evaluation at 201, 279, 290MHz && PyG-CPU, and PyG-GPU  
\\
\\
\multirow{2}{*}{\makecell{GraphACT \\ \cite{zeng2020graphact}}}
& FPGA (Xilinx Alveo U200 at 200MHz) & \makecell{\checkmark \\ (training speed)} & $12-15\times$, and $1.1-1.5\times$ over \\ & on-board evaluation && CPU, and GPU
\\
\\
\multirow{2}{*}{\makecell{Zhang et al. \\ ~\cite{zhang2020hardware}}}
& FPGA (Xilinx Alveo U200 at 250MHz) & \xmark & $30\times$, and $2\times$ over\\
& on-board evaluation && CPU, and GPU - TensorFlow and C++ (Cuda)
\\
\\
\multirow{2}{*}{\makecell{BoostGCN \\ \cite{zhang2021boostgcn}}}
& FPGA (Intel Stratix 10 GX 10M at 250MHz) & \xmark& $100\times$, $30\times$, $3$-$45\times$ over CPU (PyG $\&$ DGL), \\
& on-board evaluation &&  GPU (PyG $\&$ DGL), and FPGA (\cite{zhang2020hardware} $\&$ HyGCN)
\\
\\
\multirow{2}{*}{\makecell{G-CoS \\ ~\cite{zhang2021g}}}
& \multirow{2}{*}{\makecell{FPGA (Xilinx VCU128 at 330 MHz)\\ on-board evaluation}} & \xmark & $5.52\times$, and $1.92\times$ speedups over \\
&  && HyGCN, AWB-GCN
\\
\\
\multirow{2}{*}{\makecell{GCoD \\ ~\cite{you2022gcod}}}
& \multirow{2}{*}{\makecell{FPGA (Xilinx VCU128 at 330 MHz)\\ on-board evaluation}} & \xmark & \multirow{2}{*}{\makecell{$7.8\times$, and $2.5\times$ over\\ HyGCN, AWB-GCN}} 
\\
\\
\\
\multirow{2}{*}{\makecell{I-GCN \\ ~\cite{geng2021gcn}}}
& \multirow{2}{*}{\makecell{FPGA (Intel Stratix 10 SX at 330 MHz)\\on-board evaluation}} & \xmark & \multirow{2}{*}{\makecell{$368\times$, $453\times$, $16\times$, and $5.7\times$ over \\ PyG-GPU,  DGL-GPU, SIGMA, and AWB-GCN}} \\
\\
\bottomrule
\end{tabular}\label{tab:hw_performance}
\vskip -0.1in
\end{table*}

Table~\ref{tbl:hw_perf} reviews accelerators, focusing on validation methodology and reported speedup. These accelerators have been implemented on FPGAs or simulated as ASIC designs. A prevalent baseline is CPU and GPU, with most works utilizing PyG. Reported results may vary regarding whether they exclusively measure main kernel runtime or include other overheads for end-to-end runtime.

\subsection{GNN Accelerator Design}

Designing a GNN accelerator requires balancing computation and memory resources, since GNN workloads are typically \textit{memory-bound}. Sparse adjacency matrices lead to irregular memory accesses and high data movement costs, making dataflow and memory hierarchy optimization important. An effective design achieves balance by combining:
\begin{itemize}[leftmargin=2em]
    \item Workload preprocessing and partitioning: to transform irregular sparse operations into more regular chunks, improving mapping to hardware components and increasing efficiency.
    \item A memory hierarchy and dataflow: to minimize off-chip transfers, increase on-chip data reuse, and exploit caching and compression strategies.
    \item Compute parallelism: to saturate available on-chip bandwidth without wasting cycles.
\end{itemize}
Customized GNN accelerators range from unified array designs for general GNNs (e.g., EnGN's single PE array with ring-based aggregation) to hybrid/tiled pipelines with dedicated resources per stage (HyGCN), and tightly specialized or template-based designs customized for target applications (AWB-GCN and BoostGCN). Each balances compute and memory differently, reflecting trade-offs between flexibility, efficiency, and target workload characteristics. Key differences include:
\begin{itemize}[leftmargin=2em]
    \item On-chip buffer vs parallel compute (processing elements, i.e., PEs):
    More on-chip memory enables larger tiles (partitions) of the graph to be stored and processed locally, at the expense of making the design more expensive. On the other hand, scaling up the number of PEs without sufficient local memory risks underutilization, since PEs will stall waiting for data. The optimal balance depends on the graph and network structure. In GNNs with large feature vectors, adding more PEs is beneficial since the compute cost grows with $O(f_{in} * f_{out})$. Otherwise, the design can be memory-bound, and more on-chip resources are more beneficial.
    \item Memory bandwidth vs compute intensity: GNN aggregation has low arithmetic intensity, and adding PEs without increasing memory bandwidth may yield limited gains. To achieve meaningful speedups, accelerators must prioritize data movement efficiency. Dataflow optimization includes incorporating high-bandwidth memory and designing intelligent memory subsystems such as degree-aware caches. Emerging hardware paradigms for in-memory computing fundamentally address it through Processing-In-Memory (PIM) and In-Memory Computing (IMC) architectures. These designs minimize data movement by performing computation directly where data resides. Recent directions include GraphR~\cite{song2018graphr} and ReGNN~\cite{liu2022regnn} utilizing Resistive RAM (ReRAM) crossbar arrays to perform MVM, eliminating the need to fetch edge weights from memory.
    More recent works address challenges of graph irregularity and hardware reliability. For instance, CoDG-ReRAM~\cite{luo2022codg} employs semi-structured pruning to align graph sparsity with crossbar structures, while FARe~\cite{dhingra2024fare} introduces fault-aware training to mitigate analog noise inherent in ReRAM devices.
    Platforms like Newton~\cite{He2020NewtonAD} and GNNear~\cite{Zhou2021GNNearAF} integrate digital compute units directly within HBM banks. This allows element-wise aggregation to be offloaded to the memory, as demonstrated by commercial implementations like Samsung HBM-PIM~\cite{lee2021hardware} and software frameworks for real PIM hardware like PyGim~\cite{giannoula2024pygim}.    
    Additional optimizations like index and feature compression and operation fusion/pipelining help ensure that once a vector or edge is fetched, it is maximally reused before being evicted. 
    \item Sparse vs. dense engines: GNNs combine sparse operations (e.g., aggregation) and dense operations (e.g., feature transformation). Some accelerators integrate dedicated sparse and dense units, while others attempt to reuse the same hardware for both. The choice depends on graph sparsity and feature dimensions. Applications dominated by dense operations may benefit from unified compute arrays, as more resources can be allocated for them. Workloads with irregular sparsity often see gains from a separate, specialized sparse engine.
    \item Workload partitioning and pre-processing: Graph clustering, reordering, or preprocessing can make workloads more regular, improving utilization of compute engines. This reduces stalls and memory waste by converting sparse patterns into denser forms, though at the cost of additional precomputation and algorithmic complexity.
    \item Customization vs. unified engines: Accelerator design choices are strongly tied to the diversity of target graphs and GNN models. Some works aim to have a general architecture that can support more varieties. This, of course, comes at the expense of lower peak performance. While others have narrowed down their application, and hence, can customize deeply to better tune the memory hierarchies and compute resources for higher performance.
\end{itemize}

\section{GNN Acceleration: Special Graphs and Special GNNs}\label{sec:method_broader}
Acceleration techniques discussed above are for regular message-passing GNNs on regular graphs. However, many practical problems are better modeled with special graphs, e.g., heterogeneous or dynamic graphs (both defined below), and solved with more sophisticated GNNs, e.g., heterogeneous GNNs and GNNs with attention. Many methods above may be applied to these special cases if their specialties are ignored for suboptimal results, e.g, treating a heterogeneous graph as homogeneous. In this section, we discuss techniques for handling these special cases.

\subsection{GNN Acceleration on Heterogeneous Graphs}
The graphs discussed so far have nodes/edges of the same type, which are called homogeneous graphs. The graphs in the real world are often associated with various types of nodes/edges~\cite{dong2020heterogeneous, Javari2020WeaklySA, SS-AGA}, resulting in heterogeneity. For example, in a citation graph, nodes can be authors, venues, and papers, and edges can be an author publishing a paper, a paper published in a venue, etc. Different types of nodes and edges have substantially different meanings, and it is better to distinguish them explicitly. Heterogeneous graphs as a new data structure are defined below.
\begin{definition} \label{def:hetero}
A heterogeneous graph is defined as a directed graph $\gG = (\V, \E)$ associated with a node type
mapping function $\phi: \V \rightarrow \A$ and an edge type mapping function
$\tau: \E \rightarrow \R$. Each node $v \in \V$ belongs to one type $\phi(v) \in \A$ and each edge $e \in \E$ belongs to one type $\tau(e) \in \R$.    
\end{definition}
Many GNNs are specially designed for heterogeneous graphs~\cite{hu2020heterogeneous, hetgnn, hgk-gnn}. They face similar scalability challenges as regular GNNs on homogeneous graphs and require acceleration. In the previous sections, we discussed acceleration methods that could be used for GNNs on heterogeneous graphs by disregarding their heterogeneity (i.e., different node and edge types). However, applying acceleration methods in a homogeneous manner by ignoring heterogeneity often leads to a significant decrease in performance. For example, for sampling algorithms, if a message passing step sample node neighbors uniformly without considering the node types, some infrequent types are likely missed out. Completely missing one type of node can downgrade the GNN performance a lot~\cite{hetgnn}. A better sampling algorithm on heterogeneous graphs should perform a weighted sampling to include neighbor nodes of all types. Similarly for model pruning, heterogeneous GNNs often have separate model weights in the update function for each edge types~\cite{rgcn2018}, and ignoring heterogeneity may prune the weights completely for some edge types, which makes the GNN unable to pass messages that type of edges and downgrade the performance. Therefore, accelerating GNNs while taking into account heterogeneity presents additional challenges and requires special treatment to capture heterogeneity. In the following, we will provide a summary of recent research on modeling and accelerating GNNs on heterogeneous graphs. Please refer to Appendix~\ref{app:broader} for detailed discussions of the representative work HetGNN~\cite{hetgnn}, HGT~\cite{hu2020heterogeneous}, and Pigeon~\cite{wu2023pigeon}.

\subsection{GNN Acceleration on Dynamic Graphs}
Spatio-temporal data appears frequently in scientific research and real-world applications~\cite{zijie2020lgode, sanchez2020learning}. Dynamic graphs have shown great performance for modeling such spatio-temporal data to solve problems like molecular dynamics in biology~\cite{yang2021deep}, information diffusion over social networks~\cite{wang2021dydiff}, and epidemic spread in public health~\cite{zijie2021cgode}. Dynamic graphs are graphs that can change over time. Changes include the addition and removal of nodes and edges to capture the evolving nature of the data, which impose challenges for learning and reasoning on dynamic graphs as opposed to static graphs. Therefore, researchers have developed special spatio-temporal GNNs to jointly model the graph structural information on the spatial aspect and the evolving dynamics on the temporal aspect. The regular message passing in Equation~(\ref{eq:message_passing}) has been modified as shown in Equation~(\ref{eq:message_passing_dynamic}) for spatio-temporal GNNs, where the node representations are updated by both their spatial neighbors and their historical states.
Here $\E^{t-1}$ denotes the edges for graph at timestamp $t-1$, and sometimes the edges can be latent and need to be inferred before message passing. 
\begin{align} 
    \vh_{v}^{t} = \update (\vh_{v}^{t-k:t-1}, \aggr (\{\phi(\vh_v^{t-1}, \vh_u^{t-1}) | (u^{t-1},v^{t-1} \in \E^{t-1} \})) \label{eq:message_passing_dynamic}
\end{align} 

Spatio-temporal GNNs are often applied to computation-heavy problems like simulating large dynamical systems~\cite{sanchez2020learning}, which have similar scalability concerns as GNNs on static graphs and require acceleration. However, spatio-temporal GNN acceleration incurs special challenges because model training needs to be done chronologically and graph operations (modification and sampling) need to consider both spatial and temporal neighbors. For example, the sampling methods introduced in Section~\ref{subsec:sampling} may be individually applied to dynamic graphs at each timestamp $t$, but they can fall short in performance as they only consider spatial properties of nodes but omit the temporal influence. Recent works on spatio-temporal GNN acceleration, TGL ~\cite{zhou2022tgl}, MTGNN~\cite{wu2020MTGNN}, GraphODEs~\cite{zijie2020lgode, zijie2021cgode}, and GN-ETA~\cite{derrow2021eta} are discussed in detail in Appendix~\ref{app:broader}.
Overall, spatio-temporal GNNs for dynamic graphs are extensions of static GNNs with special treatment for the temporal aspect. Traditional methods usually employ recurrent NNs to capture the evolving nature, but are usually time-consuming and fail to capture the long-term dependency. Recent advances to use positional encoding and neural ODEs have shown promising results in speeding up dynamic GNNs. The works discussed above accelerate spatio-temporal GNNs in various ways to scale up them to computation-heavy dynamic system modeling and web-scale applications.

\section{GNN Acceleration: Cross-Cutting Synthesis and Selection Guidelines}\label{sec:cross_cutting}
As discussed in previous sections, GNN acceleration spans a vast design space ranging from algorithmic acceleration to system-level optimizations and customized hardware. While these techniques are often developed in isolation, practical applications require a cross-cutting perspective to select the most appropriate strategy for a specific workload and constraint set. In this section, we synthesize the trade-offs and provide explicit guidance for practitioners.

\subsection{Selecting Training Acceleration Strategies} 

The choice of training acceleration depends primarily on the training regime (single-run vs. multi-run) and resource constraints (memory vs. compute). For amortization scenarios like hyperparameter tuning, graph modification methods are most advantageous when the same graph is used to train many different models or hyperparameter configurations. In these multi-run scenarios, the heavy up-front cost of preprocessing or synthesizing a condensed graph is amortized over hundreds of training runs, resulting in significant net time savings.
For single-run training on very large graphs where memory capacity is the bottleneck, sampling methods are preferable. Unlike condensation, sampling requires negligible preprocessing and dynamically breaks the dependency explosion, allowing huge graphs to be trained on limited-memory devices.
When strict convergence guarantees are required, staleness-aware pipeline systems offer a middle ground. They provide the throughput benefits of approximate methods while maintaining convergence rates closer to full-batch training compared to aggressive sampling or sparsification.

\subsection{Selecting Inference Acceleration Strategies} 

For inference acceleration, the primary constraints shift from convergence speed to latency and throughput.
For real-time applications where single-query latency is paramount, pruning and quantization are the most effective tools. These methods directly reduce the model size and arithmetic intensity. Crucially, their benefits are maximized when combined with customized accelerators. For instance, sparse-aware accelerators can skip zero-computations from pruned models, and FPGA/ASIC designs can implement low-bitwidth arithmetic that COTS GPUs cannot natively accelerate, yielding speedups an order of magnitude higher than software optimization alone.
For offline batch processing like nightly recommendation updates, distillation into ``graph-less'' student models, e.g., MLPs, offers the highest throughput. By removing the neighbor dependency entirely during inference, these methods eliminate large neighbor fetching latency and the irregular memory access bottleneck, allowing the workload to run efficiently on standard CPUs or dense GPU kernels.

\subsection{Platform Selection: COTS vs. Customized Hardware} 
Choosing the hardware platform requires balancing development cost, flexibility, and performance. COTS systems (GPUs/CPUs) are the default choice for research and rapidly evolving applications due to high programmability. Performance on COTS systems is typically memory-bound. Design choices should focus on maximizing data locality through reordering and kernel fusion. COTS systems are recommended for dynamic workloads where the graph structure or GNN architecture changes frequently, preventing over-specialization.
Transitioning to customized hardware (FPGA/ASIC) is justified when the workload is stable and extreme low latency is critical. Specifically, for GNNs dominated by sparse operations, hardware with in-memory computing capabilities or specialized sparse engines can overcome the memory wall that limits GPU performance. However, this comes at the cost of reduced flexibility. Thus, hybrid architectures that retain some programmability are recommended when occasional model updates are required.

\section{Future Directions}\label{sec:future}
While significant progress has been achieved, the investigation into GNN acceleration remains in its early stages and has not kept pace with the rapid expansion of graph data. Numerous promising research avenues exist within this field, particularly in terms of acceleration from a COTS system and a customized hardware perspective, which have only recently begun to garner attention. Additionally, there is a need to explore acceleration techniques tailored for heterogeneous and dynamic graphs. This section highlights several captivating research directions.

\textbf{Better Graph Modification Algorithms} Improving the efficiency of graph modification methods, either coarsening, sparsification, or condensation, represents a promising avenue for future research. In many graph datasets, nodes often have an excessive number of neighbors, which may provide redundant information or even noise. To accelerate GNN training, it may be possible to remove these neighbors more aggressively at the pre-processing stage. For example, through sparsification without preserving the global graph properties as practical GNNs are trained for preserving locality. Graph condensation is another promising technique that can enhance training efficiency, but it currently requires running the training process at least once to identify the best condensed graph. If the optimal graph can be determined earlier, e.g., identify a good condensed graph by training only a few epochs, this technique would become much more useful. Good graph modification methods can also benefit the automating designing and tuning GNN architectures. In practice, preprocessed simpler graphs will accelerate the whole pipeline of GNN development.

\textbf{Better Sampling Algorithms} There are still several potential improvements for future research in sampling methods for GNNs. One potential direction is to investigate hierarchical sampling methods that can exploit the multi-scale structure of graphs to improve sampling efficiency and capture multi-scale features of graphs. Hierarchical sampling can be particularly useful for large-scale graphs with complex structures, such as social networks, where nodes can be organized into clusters or communities at different levels. Another direction is to further investigate the sampling method for dynamic graphs, where nodes and edges change over time. Individually applying sampling to dynamic graphs at each timestamp is not very efficient and fails to capture the temporal dynamics.

\textbf{Combined Inference Acceleration Algorithms} Inference acceleration methods, such as pruning, quantization, and distillation, are often orthogonal to each other and can be combined to improve the speed. Some initial steps have already been taken, such as leveraging distillation when ambitiously quantizing weights to binary. As inference is more flexible training, e.g., no concerns like training stability and model convergence, these methods can be more easily combined in various ways. For instance, pruning and quantization can be combined to achieve a sparse model with low precision. Additionally, pruning can be used in conjunction with distillation to further refine the distilled student model.

\textbf{COTS Systems} Existing work mainly focuses on improving the run-time efficiency of GNNs. In contrast, the memory requirement can often be the bottleneck of accelerating GNNs on large-scale data, which has not been well studied yet. Various techniques including rematerialization, quantization, and data pruning can be considered for reducing the memory overhead. In addition, because most GNN models have irregular computation and data access patterns, the advanced training/inference algorithms can be system unfriendly. This makes system-algorithm co-design for system-friendly GNN algorithms a promising research direction. Finally, full-stack development for GNN systems is still missing in the existing literature. Combining existing techniques at different levels (e.g., algorithm, system, hardware) towards an ultra-efficient GNN system is another interesting research topic.

\textbf{Customized Hardware} As explained in Section~\ref{sec:method_hardware}, memory access presents a significant bottleneck in GNN acceleration. Thus, a key aspect of accelerating the processing of GNNs is to minimize the amount of data that must be fetched from off-chip memory. In addition to the techniques employed by the works reviewed, such as data reuse, on-chip memory caching, and dataflow optimization, exploring approaches such as pruning, quantization, and compression may reduce the volume of data fetched from off-chip memory. 
Furthermore, a critical requirement is to develop a more advanced automated design generator that facilitates GNN accelerator design. Techniques such as Neural Architecture Search (NAS)\cite{lin2021naas}, learning-based models for accelerated hardware design evaluation\cite{gnn-dse}, and intelligent design space exploration to navigate through design candidates\cite{autodse, mirhoseini2021graph} can be suitably adapted for GNN accelerator design.

\textbf{Dynamic Graphs} While recent works have made strides in scaling temporal GNNs on dynamic graphs, several critical challenges remain unresolved, marking promising directions for future research. First, most existing methods treat dynamic graphs as a sequence of snapshots or a stream of events but still perform heavy re-computation for inference or training updates. A major open challenge is developing efficient incremental update mechanisms that can adjust model states in response to graph evolutions without triggering a full-graph re-computation.
Second, current systems often underutilize the high temporal locality inherent in dynamic graphs. Future systems could design intelligent caching policies and memory hierarchies that prioritize these temporal subgraphs, minimizing data movement by keeping active contexts on-chip or in GPU memory.
Finally, standard static graph modifications and pre-processing become prohibitively expensive if re-run for every timestamp. There is a significant need for lightweight, dynamic maintenance algorithms that can update partitions and indices incrementally, amortizing these system costs over the lifespan of the streaming graph.


\bibliographystyle{ACM-Reference-Format}
\bibliography{reference}

\appendix
\clearpage
\section{Notations and Preliminaries}\label{sec:background}
In this section, we clarify the common notations used throughout the paper in Table \ref{tab:notation}. Also, we review graph machine learning with GNNs by going through the graph data representation, the GNN model architecture, and the machine learning tasks GNNs can solve.

\begin{table}[th]
\center
\footnotesize
\caption{Major Notations}
\begin{tabular*}{\textwidth}{@{}lllllll@{}}
\toprule
\multicolumn{1}{l}{Category} & Notations & Descriptions \\ \midrule
\multirow{10}{*}{Graphs}
& $\gG$ & A graph with feature-enriched nodes \\
& $N$ & The total number of nodes in $\gG$, i.e., $|\V|$ \\
& $\V$ & The node set of $\gG$ \\
& $\E$ & The edge set of $\gG$ \\
& $\mX$ & Raw node features in $\R^{N \times D}$\\
& $\vx_i$ & The $i$th row of $\mX$, which is the raw feature of the node $i \in \V $ \\
& $\mA$ & The adjacency matrix. $\mA \in \{0, 1\}^{N \times N}$. $\mA_{i,j} = 1$ if edge $(i ,j) \in \E$, otherwise 0 \\
& $\tilde \mA$ & The adjacency matrix with self-loops. $\tilde \mA = \mA + \mI$ \\
& $\mD$, $\tilde \mD$ & Degree matrices of $\mA$ and $\tilde \mA$. $\mD_{i,i} = \sum_{j}\mA_{j, i}$ and $\mD_{i,j} = 0$ if $i \neq j$. Similarly for $\tilde \mD$ \\
& $\N(v)$ & Neighbors of a node $v \in \V$. $\N(v) = \{u \in \V | (u, v) \in \E \}$ \\ 
& $d(v)$ & The degree of node $v$. $d(v) = |\N(v)|$ \\\midrule
\multirow{12}{*}{GNNs}
& $\gnn_{\theta}$ & The GNN model with parameters $\theta$ \\
& $L$ & Total number of layers of a GNN\\
& $\mH^{(l)}$ & Hidden node representations at layer $l$. $\mH^{(0)} = \mX$. $\mH^{(l)} \in \R^{N \times F}$ (assume $F$ is shared for $l \geq 1$) \\
& $\vh_i^{(l)}$ & The $i$th row of $\mH^{(l)}$, which is the hidden feature of node $i$ at layer $l$ \\
& $\phi(\cdot, \cdot)$ & The message (a.k.a. feature extraction) function which constructs a message between two nodes \\
& $\vm_{u, v}^{(l)}$ & The constructed message from node $u$ to node $v$ at layer $l$ \\
& $\aggr$ & The aggregation function, usually order invariant, e.g., mean \\
& $\va_v^{(l)}$ & The aggregated feature of node $v$ at layer $l$ \\
& $\update$ & The update function. A neural network. Usually a multi-layer perceptron (MLP) \\
& $\mW^{(l)}$ & Neural network weights need to be learned in the $\update$ function at layer $l$\\
& $\sigma$ & The nonlinear activation function of the $\update$ neural network \\
& $\mP$ & The filter matrix for aggregating messages, usually a function of $\mA$ and $\mD$ \\ \midrule
\multirow{5}{1pt}{\thead{Labels \& \\ Training}} 
& $\mY$ & One-hot node labels to predict. $\mY \in \{0, 1\}^{N \times C}$ \\
& $\vy_i$ & The $i$th row of $\mY$, which is the one-hot node label for the node $i$ \\
& $\hat\vy_i$ & Prediction of the node $i \in \V$, a probability vector. $\hat\vy_i \in [0, 1]^{C}$\\
& \onehot & One-hot vector function. Takes $c \in \{1, \dots, C\}$ and outputs $\vy \in \{0, 1\}^{C}$ with only $\vy_c = 1$\\
& $\Lo(\vy_i, \hat\vy_i)$ & The loss function for training $\gnn_{\theta}$ \\
\bottomrule
\end{tabular*}
\label{tab:notation}
\end{table}

\subsection{Graph Data Representation} \label{subsec:background_graph_data}
An attributed graph $\gG$ includes the graph structure $(\V, \E)$ and node features $\mX$, with $\V$ denotes a set of nodes, and $\E$ denotes a set of edges. A straightforward way to represent the graph structure (edges $\E$) is to use an adjacency matrix $\mA \in \R^{|\V| \times |\V|}$, with $|\V|$ denotes the number of nodes. Many graph algorithms, including the matrix version of GNNs, are presented in terms of $\mA$. However, naively representing edges with the full matrix $\mA$ requires $O(|\V|^2)$ space complexity, which is impractical to implement for large graphs. In practice, most of the graphs are sparse, i.e., $|\E| << |\V|^2$. Leveraging the sparsity to represent $\gG$ in more efficient formats can help save memory and accelerate many computations. One popular sparse representation is the \textit{Coordinate (COO)}~\cite{wang2019dgl} format, which is also known as the ``triplet'' format. For this format, each edge $(i, j) \in \E$ is represented as a triplet $(i, j, w_{ij})$ with $w_{ij}$ being the edge weight for weighted graphs, and $w_{ij}$ equals one for unweighted graphs. The space complexity of a COO graph scales only with the number of edges, e.g., $O(|\E|)$. Another popular choice is the \textit{Compressed Sparse Row (CSR)} format~\cite{CSR, wang2019dgl}. It is similar to the COO format but compresses those $|\E|$ triplets into three $|\E|$-dimensional arrays. One array for all the row indices, e.g., the first entry in COO. One array for all the column indices, e.g., the second entry in COO. One array for all the edge weights, e.g., the third entry in COO. The space complexity of a CSR graph is also $O(|\E|)$. 
Additionally, graphs often come with node features, which are represented as a feature matrix $\mX \in \R^{|\V| \times D}$. $\mX$ is often referred to as the raw features to distinguish them from the hidden features learned by the model. Similarly, there could be features for graph edges, but we do not assume edge features are given in this survey because they are not generally available.

\subsection{Graph Neural Networks} \label{subsec:background_gnn}
GNN is a family of neural network models that achieve state-of-the-art performance on many machine learning tasks on graphs. GNNs take a graph and node features as input and output node representations. The node representations can be used for various downstream ML tasks like node classification and edge prediction. Node representations can also be aggregated over the whole graph to get a graph representation for graph classification.

There are many different GNN model architectures. For example, the graph convolutional network (GCN)~\cite{gcn}, graph information network (GIN)~\cite{ginconv2018}, graph attention network (GAT)~\cite{gat}, etc. Most of the model variants belong to the \textit{message-passing} GNN framework~\cite{mpnn}. Message passing is a mechanism that iteratively updates the hidden representation of a node by aggregating messages from its neighbors and transforming the aggregated messages with a neural network. Each message-passing iteration is called a GNN layer, which allows the hidden representation to combine information from the node's direct neighbors. Stacking multiple GNN layers allows multi-hop neighbors to pass messages to the central node. Thus, the node representation can capture the contextual information of a local neighborhood and be used to answer complex ML questions on graphs. 

We now describe the detailed message-passing operation using the notations defined in Table \ref{tab:notation}. For layer $l$, the hidden representation $\vh^{(l)}$ of a node $v$ is updated by first aggregating messages (a function of hidden representations from the previous iteration) from its neighbors and then combining the aggregated messages with its own hidden representation. The message-passing operation: 
\begin{align} 
\vh_{v}^{(l)} = \update (\vh_{v}^{(l-1)}, \aggr (\{\phi(\vh_v^{(l-1)}, \vh_u^{(l-1)}) | u \in \N(v) \})). \label{eq:message_passing}
\end{align} 
where $\phi(\cdot, \cdot)$ constructs the message between two nodes from their hidden representations; \aggr{} is the aggregation function, usually being order-invariant, e.g., mean; and  \update{} is the update function, usually being a multi-layer perceptron (MLP) with learnable weights $\mW^{(l)}$. Equation (\ref{eq:message_passing}) shows one message passing step for layer $l$, i.e., the $l$-th GNN layer. For a GNN with $L$ total layers, message passing is applied $L$ times to aggregate messages from neighbors within $L$-hops of $v$ to compute the representation $\vh^{(L)}_v$. All the neighbors within $L$-hop of $v$ form the \textit{receptive field} of $v$. A recursively constructed $L$-layer tree graph with $v$ as the root and the neighbors of each node as its children are called a \textit{computation graph} of $v$.

For many GNNs, the message-passing operation in Equation (\ref{eq:message_passing}) can be implemented by a clean matrix multiplication to update hidden representations of all $v \in \V$ at once, which is also called the \textit{full-batch} version. The full-batch updates:
\begin{align} \label{eq:message_passing_matrix}
\mH^{(l)} &= \sigma (\mP \mH^{(l - 1)} \mW^{(l)}),
\end{align} 
where the matrix $\mH^{(l - 1)}$ contains hidden representations of all nodes at the $(l-1)$-th step. It is first multiplied by a \textit{filter matrix} $\mP$ (on the left) and a neural network weight matrix $\mW^{(l)}$ (on the right) and then is passed through a nonlinear activation function $\sigma(\cdot)$. $\mP$ corresponds to the \aggr{} function and $\mW^{(l)}$ and $\sigma(\cdot)$ correspond to the \update{} function (when it is an MLP) in Equation (\ref{eq:message_passing}). $\mP$ can take different forms for different GNNs, and it is usually a function of $\mA$. For example, $\mP$ is a degree-normalized version of the adjacency matrix with self-loops for GCN, i.e., $\mP = \tilde \mD ^{-\frac{1}{2}} \tilde \mA \tilde \mD ^{-\frac{1}{2}}$. 

\begin{algorithm}[hbt]
  \caption{SAGA (Scatter-ApplyEdge-Gather-ApplyVertex) based GNN Message Passing}
  \label{alg:message_passing}
\begin{algorithmic}
  \STATE {\bfseries Input:} A graph $\gG = (\V, \E)$, node features $\mX$
  \STATE $\mH^{(0)}= \mX$
  \FOR{$l \gets$ 1 to $L$}
    \FOR{(u, v) in $\E$}
        \STATE $\vm_{u, v}^{(l)} \gets \phi(\vh_v^{(l-1)}, \vh_u^{(l-1)})$
    \ENDFOR
      \FOR{v in $\V$}
        \STATE $\va_{v}^{(l)} \gets \aggr(\{\vm_{u, v}^{(l)} | u \in \N(v) \})$
        \STATE $\vh_{v}^{(l)} \gets \update (\vh_{v}^{(l-1)}, \va_{v}^{(l)})$
        \ENDFOR
    \ENDFOR
\end{algorithmic}
\end{algorithm}

The matrix version looks cleaner but it is implemented less often in practice since the full matrix representation of $\mP$ usually has space complexity $O(|\V|^2)$, so it is infeasible to fit such $\mP$ in memory for large graphs. More frequently,  $\mP$ is represented as a sparse matrix (e.g., in COO or CSR format) mentioned above, and message passing is implemented using the Scatter-ApplyEdge-Gather-ApplyVertex (SAGA) paradigm~\cite{saga}. For SAGA, messages are first computed and stored on edges, and then each node aggregates messages from its edges and updates the representation. We show the SAGA GNN message passing in Algorithm \ref{alg:message_passing}, which has time complexity $O(|\E|)$.

\subsection{Graph Machine Learning Tasks}
GNNs are capable of solving many graph machine-learning tasks. For example, at the node level, it can classify fraudulent nodes. At the edge level, it can predict edges for recommendations. At the graph level, it can classify properties for molecule graphs. For all three kinds of tasks, the standard approaches share a common step of getting node representations $\mH$. Then $\vh_{v}$ can be used for solving node-level tasks, or pairs $(\vh_{u}, \vh_{v})$ can be used for solving edge-level tasks, and $\mH$ may go through a pooling layer over all of its rows to become a single representation for solving graph-level tasks. The last prediction step is usually running through a few fully-connected network layers and is not time-consuming \cite{gcn}. Therefore, for GNN acceleration, we focus on how $\mH$ can be efficiently learned on large graphs, and we use multi-class node classification as a concrete example in this survey.

Node classification is one of the most widely considered tasks on graphs, which classifies each node $v \in \V$ to one of $C$ classes. It is usually semi-supervised meaning that the entire graph with all node features and a few node labels are observed. The goal is to leverage the observed labels and the graph information to classify other nodes without labels. When inferring the class of node $v$, its final layer hidden representation $\vh_{v}^{(L)}$ is first generated via message passing, and then $\vh_{v}^{(L)}$ is plugged into a prediction head $f: \R^F \rightarrow \R^C$ to get the final prediction $\hat \vy_{v} = f(\vh_{v}^{(L)})$. $f(\cdot)$ is usually an MLP learned jointly with the GNN using the cross-entropy loss. The loss will be computed for all nodes with observed labels during training, and for a single node $v$, it is:

\begin{align} \label{eq:loss} 
\Lo (\vy_{v}, \hat \vy_{v}) = \sum_{c=1}^{C} \vy_{v, c} \log \hat \vy_{v,c}.
\end{align}

\section{Detailed Discussion of Training Algorithms} \label{app:alg_training}

\subsection{Graph modification}
\textbf{Huang et al.}~\cite{scale_up} discuss a general framework using graph coarsening to accelerate GNN training. In particular, a graph $\gG$ is partitioned into K clusters with sizes $N_1, ..., N_K$. A matrix $\mC \in \R^{N \times K}$ is used to represent the partition, where $\mC_{i,j} = 1/\sqrt{N_j}$ if and only if node $i$ is assigned to cluster $j$, and $\mC_{i,j} = 0$ otherwise, so that $\mC^{\top}\mC = \mI$. The super-nodes and super-edges are constructed according to $\mC$. Moreover, the super-node features $\mX'$ are set to be the weighted average of node features within each cluster, i.e., $\mX' = \mC^{\top} \mX$. The super-nodes labels $\mY'$ are constructed by taking the dominant label of nodes in each super-node, i.e., $\vy'_{k} = \onehot(\argmax ((\mC^{\top} \mY)_{k}))$. \cite{scale_up} considers different partition algorithms for coarsening the graph: spectral clustering \cite{spectral_clustering} and several partition algorithms from \cite{loukas_coarsening2019} with restricted spectral approximation. Empirically, the Variation Neighborhoods method from \cite{loukas_coarsening2019} works better than other partition algorithms. 

\subsection{Sampling}

\subsubsection{Benchmark Datasets}

We compared popular sampling methods in runtime and accuracy in Table \ref{tab: sampling performance}. The results are on three commonly used benchmark datasets: Pubmed \cite{cora}, PPI \cite{zitnik2017predicting}, and Reddit \cite{hamilton2017inductive}. \textit{Pubmed}~\cite{cora} is a citation graph of 19,717 scientific publications on diabetes from the PubMed database as nodes and 44,338 citations as links. Each publication has a TF/IDF word vector feature~\cite{manning2009introduction}. The task is to classify nodes into one of three diabetes types. \textit{PPI}~\cite{zitnik2017predicting} is a protein-protein interaction graph with 14,755 nodes and 225,270 edges, where each node has features based on positional gene sets, motif gene sets, and immunological signatures. The task is to classify nodes into one of 121 genes. \textit{Reddit}~\cite{hamilton2017inductive} is a graph constructed from Reddit posts with 232,965 nodes (posts) and 11,606,919 links (connections when the same user comments on both posts). Node features are vectors concatenating the average embedding of the post title, comments, score, and number of comments, using GloVe CommonCrawl word vectors~\cite{pennington2014glove}. The task is to classify posts into one of 41 Reddit communities.

\subsubsection{Algorithm Details}

\textbf{AS-GCN} \citep{huang2018adaptive} proposes an adaptive layer-wise sampling method to accelerate the training of GCNs on large-scale graphs. The proposed method samples the lower layer conditioned on the top one, where the sampled neighborhoods are shared by different parent nodes and over-expansion is avoided. AS-GCN explicitly calculate the embedding variance, and add a regularization to reduce the variance. In this way, the model could learns to give higher weights to those more connected nodes, which has been demonstrated to achieve higher accuracy than FastGCN. Additionally, AS-GCN preserves second-order proximity by formulating a skip connection across two layers.

\textbf{LADIES} \citep{zou2019layer} adopts similar intuition as AS-GCN to only sample within the the immediate neighborhood of the later layer. This constraint can guarantee that each sampled node in the previous layer will have at least one successor node in the later layer. In other words, it can make sure all the sampled nodes can deliver their information to at least one of their neighbor nodes in the next layer. To apply this constraint, a row selection matrix $\mQ^{(l)}\in\R^{s_{layer}^{(l)}\times |\V^{(l)}|}$ is constructed as
\begin{align}
    \mQ^{(l)}_{v,u} = \left\{
 \begin{array}{ll}
    1 & \text{if } v \text{ is sampled in layer $l$ and } u\in\mathcal{N}(v) \\
    0,& \text{otherwise},
    \end{array}
\right.
\end{align}
Then, node $u$ in layer $(l-1)$ is sampled with probability: $p^{(l-1)}_{\mathcal{V}}(u) = \frac{\|\mQ^{(l)}\mP_{:,u}\|^2_2 }{ \|\mQ^{(l)}\mP \|^2_F}$. A $\overline{\mP}^{(l)}$ can be constructed in the same way as in Equation (\ref{eq: P for Layer-wise sampling}). 

To maintain the scale of embeddings in the forward process, LADIES also normalizes $\overline{\mP}^{(l)}$ to compute the $\overline{\mP}^{(l)} \mH^{(l - 1)}$. It computes $\overline{\mP}^{(l)} \leftarrow \mD_{\overline{\mP}^{(l)}}^{-1}\overline{\mP}^{(l)}$, where $\mD_{\overline{\mP}^{(l)}}$ is a diagonal matrix with its $(i,i)$ entry equals $\sum_j \overline{\mP}^{(l)}_{ij}$.

\textbf{VR-GCN} \citep{chen2017stochastic} proposes to use historical embeddings in GraphSAGE workflow to reduce embedding variance and accelerate training convergence. VR-GCN applies the same sampling operation with the same uniform probability distribution as in GraphSAGE. Under the assumption that the model weight $\mW^{(l)}$ would not change significantly between two training iterations, VR-GCN stores the historical embeddings $\mH^{(l, \ history)}$ from the previous epoch. With that, $\mP^{(l)}\mH^{(l-1)}$ is be estimated by 
$\overline{\mP}^{(l)}\mH^{(l-1)}  + \mP^{(l)}\mH^{(l-1, \ history)}$. Such historical embedding usage causes some extra memory, but it is usually acceptable. The benefit is that VR-GCN can enjoy the advantage of a smaller training variance that accelerates training convergence. In practice, \cite{chen2017stochastic} found that VR-GCN can get a comparable performance of GraphSAGE and other existing methods with a much smaller sample size.

\textbf{GraphSAINT} \citep{zeng2019graphsaint} is another widely used subgraph-wise sampling method, which first samples nodes and then constructs subgraphs induced by the sampled nodes. GraphSAINT proposes four different algorithms to sample nodes: 1) node sampler, which randomly samples nodes with probability proportional to node degrees; 2) edge sampler, which samples edges with probability proportional to the inverse of the end node degrees, and then keeps the end nodes. The edge sampler is proven to reduce the estimation variance; 3) random walk sampler, which conducts random walks on a random set of nodes and selects nodes on the walks; 4) multi-dimensional random walk, which is similar to random walk but adds an additional requirement to start random walking from high-degree nodes in the sampled set. Among the four proposed algorithms, the random-walk sampler performs better empirically and thus is more widely used. 

\textbf{Shadow-GNN} \citep{zeng2021decoupling} samples different subgraphs for each target node (similar to ego-networks). The sampling could be done via random walk or personalized page rank. Specifically, the authors introduce a design principle that separates the depth of a GNN from its receptive field size, which is the range of nodes and edges that influence a node's representation. By doing so, the depth of a GNN can be increased without increasing its receptive field size exponentially. It also presents a theoretical analysis of expressivity from three different perspectives, and also rich design components (e.g., subgraph extraction functions, architecture extensions) to implement such design principles.

\section{Detailed Discussion of Inference Algorithms} \label{app:alg_inference}

\subsection{Pruning}

One recent popular line of pruning research is the Lottery Ticket Hypothesis (LTH) \cite{frankle2018lottery}, which shows that a pruned sparse NN can be \textit{retrained} to achieve similar accuracy as the original NN. The significance of LTH compared to the regular pruning methods is in the retraining. Most pruning methods result in \textit{one} set of NN weights with minor accuracy loss, but they cannot guarantee that the pruned NN can be retrained with the original ML objective and still maintain similar accuracy loss on the same dataset. LTH shows that the same level of accuracy loss can be achieved with iterative magnitude pruning (IMP), i.e., prune the small NN weights iteratively and perform retraining after each pruning. 

\textbf{Chen et al.} \cite{chen2021unified} test LTH for GNNs by doing IMP. They indeed observe speed gain and marginal accuracy loss even for retrained GNNs. \textbf{You et al.} \cite{you2021early} further test the early-bird LTH on GNNs, which means the retraining in each IMP iteration can stop very early before it converges. Notice that pruning accelerates model inference but not always model training. The iterative pruning requires retraining the model multiple times, possibly resulting in a slower training procedure than before. The advantage of early-bird LTH is to save iterative training time for finding the optimal pruned model. Moreover, in \cite{chen2021unified} and \cite{you2021early}, the input graph is pruned together with the GNN model, so parts of these methods are doing graph pruning (sparsification) as well.  

\subsection{Quantization}

\textbf{SGQuant} \cite{feng2020sgquant} is one of the earliest work studying GNN quantization. It is a PTQ method and focuses on the two core PTQ questions mentioned above. For the first question about the quantization object, SGQuant focuses on the quantization of node representations learned by each GNN layer, i.e., $\{\mH^{(l)}\}_{1 \leq l \leq L}$. SGQuant makes the empirical observation that node representations consume much more memory than GNN weights $\mW$. This observation and object choice is unique to GNNs but not for other NNs because of node dependency. Unlike other NNs which process one data instance one at a time, for one target node, GNNs store node representations for all the multi-hop neighboring nodes within the receptive field of the target, which could involve a huge number of nodes. For the second question, SGQaunt focuses on selecting the quantization bit $q$ and simply uses the full range of $q$-representable numbers. SGQuant proposes multi-granularity quantization, where different objects are associated with different bits to optimize quantization efficiency. Applying this principle to graph data, SGQuant uses topology-aware and layer-aware quantization to group node representations, where nodes with similar degrees are grouped together, and representations from the same layer are grouped together. This is essentially grouping rows of $\{\mH^{(l)}\}_{1 \leq l \leq L}$, where each row corresponds to one node and $l$ corresponds to layer depth. Then node representations are assigned with different bits depending on which group they are from. The multi-granularity quantization decides the bits for each group by 
optimizing quantized model accuracy over a set of precision candidates. 

\textbf{Degree-Quant} \cite{tailor2020degree} is a QAT method quantizes both GNN weights $\mW$ and intermediate node representations $\mH$, which results in INT8 GNNs with comparable accuracy as their FP32 counterparts but up to 4.7X faster. Degree-Quant is like many other QAT methods for general NNs, for which a quantized model is trained via gradients by simulating its numerical errors from the original model. During the backward pass, the straight-through estimator (STE) \cite{bengio2013estimating} is used to compute gradients since the quantization operation, e.g., round to integers, can be non-differentiable. The unique challenge for GNN quantization via STE is that the STE performance is degraded for high-degree nodes, which leads to poor weight updates. Degree-Quant verifies the degradation both theoretically and empirically on common GNNs architectures including GCN, GIN and GAT. It then proposes a protective mask to improve weight update accuracy. A protective mask $\bm{\beta}$ is a length-$N$ binary vector for all $N$ nodes in graph $\gG$. During the quantization-aware GNN training, node representation $\vh_i$ with $\bm{\beta}_i = 1$ is kept with full precision, while $\vh_i$ with $\bm{\beta}_i = 0$ and all weights $\mW$ are quantized. The protective mask $\bm{\beta}$ is sampled as $\bm{\beta} \sim Bernoulli(\vp)$ with $\vp$ containing probabilities for each node. Degree-Quant treats the highest probability and lowest probability in $\vp$ as two hyperparameters, and probabilities are assigned to nodes by ranking nodes according to their degrees and interpolating these two values according to the ranking. High-degree nodes are assigned higher probabilities, which are encouraged to be kept by the protective mask for more accurate gradient updates. Degree-Quant also proposes to clip the 0.1\% top and bottom quantized values to improve model accuracy, because large fluctuations in the variance of message-passing aggregation are observed.

\subsection{Distillation}

\textbf{TinyGNN} \cite{tinygnn_kdd20} shares a similar intuition as LSP to preserve local structures, but it chooses to add a special layer termed a Peer-Aware Module (PAM) to the student model instead of defining a new KD objective as in LSP. TinyGNN adopts the standard KL divergence loss $\mathcal{L}_{i}^{KD}$. The teacher model TinyGNN considers is GAT. The student model, however, is a GAT with fewer message-passing layers than the teacher but a special PAM layer inserted before each message-passing layer. The idea of PAM is to encourage nodes in the same message-passing hierarchy to directly interact. For example, when the GNN is only one layer, a message passing step centered at node $v$ will aggregate messages from its neighbors $\N(v)$, but nodes in $\N(v)$ will not directly interact with each other during this message passing step. In other words, the computation graph is a one-layer tree and has no edges between its leaves. TinyGNN argues that sharing messages between nodes in $\N(v)$ is important for capturing the local structure around $v$. This message sharing happens naturally when the GNN is deep like the teacher, but it is lacking if the GNN is only one or two layers like the student in TinyGNN. Thus, the PAM is proposed to encourage this message sharing by adding a self-attention layer between all nodes in $\N(v)$. It was shown that a GAT student with PAM has a higher distilled model accuracy than a regular GAT student, though the speed gain is not as much given the extra self-attention computation.

We briefly mention more KD on graphs works and note that they are less related to acceleration. \textbf{GNN Self-Distillation (GNN-SD)} \cite{chen2021self} considers teacher-free self-distillation where the knowledge is extracted and transferred between layers of the same GNN. GNN-SD is shown to consistently achieve better accuracy than the standard KD objective $\mathcal{L}_{i}^{KD}$. \textbf{Combination of Parameterized label propagation and Feature transformation (CPF)}~\cite{cpf} combines KD and label propagation (LP) together. The student in CPF includes an MLP module for learning feature information and an LP module for learning structure information. The classification accuracy of a CPF student can generally outperform the accuracy of teacher GNNs after KD. However, since the LP module still has node dependency like the message passing (labels instead of the hidden features are passed as messages), the speed gain is likely limited and not discussed in \cite{cpf}. \textbf{Cold Brew}~\cite{zheng2021cold} utilizes KD to improve GNN performance on nodes with incomplete or missing neighbors, i.e., the cold-start nodes. Cold Brew greatly improves the accuracy on these cold-start nodes, but its inference stage requires a scan through the whole graph to select existing nodes as neighbor candidates for the cold-start nodes. Cold Brew thus has even more node dependency and runs slower than a standard message-passing GNN. More related works include \textbf{Graph-Free Knowledge Distillation (GFKD)}~\cite{deng2021graphfree} transfers knowledge from a teacher GNN to a student GNN via generating fake graphs. GFKD focused on graph-level prediction tasks, for which data instances are independent graphs and the scalability problem caused by node dependence is less urgent. 

\section{Detailed Discussion of COTS Systems} \label{app:system}

\subsection{GPU Kernel Acceleration}


\textbf{PCGCN}~\citep{tian2020pcgcn} improves the data locality in GNN computation by leveraging the unique sparsity pattern in graphs that nodes are usually clustered. As such, processing GNN workloads in terms of subgraphs substantially enhances the data locality. Furthermore, PCGCN employs a dual-mode computation so that SpMM (sparse matrix-matrix multiplication) is used for sparse graphs while GeMM (general matrix-matrix multiplication) is leveraged for dense graphs. For ensuring coalesced memory accesses of features, SpMM processes all edges linking each pair of partitions at each stage. The proposed technique has been implemented on top of TensorFlow. However, PCGCN relies on graph clustering, which may incur non-trivial preprocessing overhead needed for graph partition. 



\textbf{fuseGNN}~\citep{chen2020fusegnn} develops two workload abstractions of GNN aggregation, assuming that the input graph adopts the COO format: Gather-ApplyEdge-Scatter (GAS) and Gather-ApplyEdge-Reduce (GAR). Specifically, GAS works with the COO format graph, and processes edge-wise aggregation; GAR handles node-wise aggregation, which is particularly useful for the CSR format as we defined in Section~\ref{subsec:background_graph_data}. Although GAR is usually efficient because it relies on on-chip reduction with a shared output memory, creating CSR from COO is usually time costly, thus degrading the achievable efficiency of GNN computation. As a result, fuseGNN uses GAS for graphs with a low average degree (e.g., Cora\cite{cora}) and GAR for graphs with a high average degree (e.g., Reddit~\cite{sage}). All operations in the three kernels, including graph processing for graph format conversion, GAS, and GAR, are fused for alleviating the memory traffic. 
fuseGNN is developed for PyTorch and can be extended to multi-GPU scenarios.



\textbf{FusedMM}~\citep{rahman2021fusedmm} is a CPU kernel but the authors claim that it can be extended to GPU acceleration. The work divides popular graph operators into five types of basic operations that can be launched sequentially (e.g., element-wise additive, reduction operation, element-wise self-operation) and defined by users for designing customized GNN layers. During GNN computations, the operations associated with each target node are fused together for achieving memory reduction by eliminating the necessity of saving intermediate results. Since each CPU thread is assigned multiple target nodes, FusedMM further adopts an edge-balanced partition to ensure a balanced workload among threads. The proposed kernel of FusedMM can be integrated into DGL~\cite{wang2019dgl}.

\textbf{TLPGNN}~\citep{fu2022tlpgnn} develops two levels of parallelism:  node- and feature-level parallelism. In the first level (i.e., warp-level), TLPGNN assigns each target node to a warp of threads for avoiding branch divergence caused by uneven node degrees. For computing each node's new embedding, the second level (i.e., thread-level) adopts feature-level parallelism that  the threads in the same warp process feature in parallel while processing each edge sequentially, which allows coalesced memory accesses to input features for each source node. To achieve a balanced workload, a dynamic workload assignment is developed, which allocates the next computation task once one hardware resource is released. TLPGNN further performs kernel fusion to avoid unnecessary memory accesses.

\textbf{Zhang et al.}~\cite{zhang2022understanding} first decompose a GNN operation into four parts (i.e., scatter across edges, apply edge, gather from edges, and apply node), and then identify three challenges in GNN computations: redundant neural operator computation, inconsistent thread mapping, and excessive intermediate data. Specifically, redundant neural operators can be caused when the same feature transformation is employed to the same node features after they are scattered across edges. To address this issue, the scatter operation can be postponed after the feature transformation is finished. For the second issue, if the thread mappings are inconsistent (e.g., scatter threads are assigned with edge-centric tasks, while gather threads are assigned with node-centric tasks), these threads have to be processed sequentially and cannot be merged. This work leverages unified thread mapping for fusion. The third issue identifies that intermediate results may occupy up to 92\% of the total memory requirement, which can be resolved by recomputation during backward propagation.

\textbf{MaxK-GNN}~\citep{peng2024maxk} accelerates GNN training on GPUs by addressing sparsity, irregularity, and memory inefficiencies in sparse matrix-matrix multiplication (SpMM) operations. It introduces a MaxK nonlinearity that induces controlled sparsity by selecting the top-k elements per node embedding, reducing effective feature dimensions with minimal accuracy loss while preserving universal approximation capabilities. To handle the resulting sparse features, it employs a Compressed Balanced Sparse Row (CBSR) format for contiguous storage and coalesced memory access. Custom kernels are designed, including a coalescing-enhanced forward SpGEMM (sparse general matrix-matrix multiplication) with shared memory-based accumulation and warp-level partitioning for workload balance, as well as an optimized backward SSpMM (sampled SpMM) with dense row prefetching and atomic global memory accumulation. These optimizations reduce global memory traffic significantly (up to 90\%), improve cache hit rates, and enhance bandwidth utilization, leading to substantial speedups over baselines like cuSPARSE while maintaining comparable model accuracy.

\subsection{User-defined Function Optimization}
\textbf{FeatGraph}~\citep{hu2020featgraph} provides a flexible programming interface on top of the compiler stack for DL systems: TVM \cite{chen2018tvm} for supporting friendly user-defined functions. The users can manually select the tiling factor (the number of features being processed in each thread) for maximizing the cache utilization. However, because TVM lacks the support for sparse operations, FeatGraph only supports SpMM/SDDMM-based operations.

\textbf{Seastar}~\citep{wu2021seastar} develops a vertex-centric programming interface for user-defined functions, and generates an execution plan that optimizes memory consumption and data locality, resulting in higher performance. Specifically, Seastar first translates the vertex-centric logic into tensor operations via a tracer, and then identifies operator fusion opportunities by the notion of Seastar computation pattern. This approach reduces memory consumption and enhances data locality via fusing operations of each target node by leveraging the merits of vertex-centric programming, similar to FusedMM~\cite{rahman2021fusedmm}. Additionally, Seastar uses a distributed graph communication library (DGCL) to enable scalable GNN training on distributed GPUs, and its DL backend is MindSpore, an all-scenario AI computing framework \cite{mindspore}.

\textbf{Graphiler}~\citep{xie2022graphiler} develops an automated tool for scatter/gather-related tasks. Specifically, compared with \cite{zhang2022understanding} mentioned in the previous section that manually implements operation reordering for some specific GNN operations, Graphiler enjoys better flexibility by automatically detecting redundant communication and then adjusting their order accordingly, and further fuses all operations after gathering from edges to reduce intermediate memory accesses.

\textbf{gSampler}~\citep{gong2023gsampler} introduces a general and efficient GPU-based framework for graph sampling in graph learning tasks. It proposes the Extract-Compute-Select-Finalize (ECSF) programming model to unify a wide range of graph sampling algorithms, allowing users to express complex sampling logic through matrix-centric APIs. gSampler optimizes execution via a data flow graph intermediate representation, incorporating computation optimizations to eliminate redundant data movements, data layout optimizations to select efficient storage formats based on algorithms and datasets, and super-batching to enhance GPU utilization without affecting accuracy. Compared to systems like DGL, gSampler achieves average speedups of 6.54× in sampling, reducing overall GNN training time by over 40\%.

\subsection{On-device scalable training systems}

\textbf{CAGNET}~\cite{tripathy2020reducing} develops four communication patterns by partitioning the adjacency matrix of target GNNs into 1D, 1.5D, 2D and 3D formats and then distributing the workload across workers accordingly. This work further analyzes the communication latency for each design. For ensuring a balanced workload, the random partition is adopted. However, because communication does not consider the sparse pattern, broadcasting all features causes redundant data transfer (i.e., workers receive node features that are not required), leading to both significant communication overhead and memory consumption.



\textbf{Dorylus}~\cite{thorpe2021dorylus} is a serverless platform for saving the cost of training GNN models, enabling affordable, scalable, and accurate GNN training. Because conventional GPU servers are expensive and have restricted memory, Dorylus presents a distributed system with much lower-cost CPU clusters and Lambda threads, a representative of serverless threads. The authors notice that the training time of GNNs is dominated by neighbor propagation, rather than neural network operations. As parallel processing does not effectively enhance the speed of neighbor propagation, Dorylus executes 1) the neighbor propagation in CPU clusters for achieving the best performance per dollar, and 2) the neural network operations in Lambda threads for avoiding the necessity of unneeded resources (e.g., storage) that conventional distributed platforms (e.g., CPU clusters) require. To further minimize the network latency between the aforementioned CPU clusters and Lambda threads, Dorylus adopts fine-grained computation and asynchronous execution through pipelined computation and communication including embedding transfer and model synchronization. Following VR-GCN (see Section~\ref{subsec:sampling}), the convergence analysis based on historical embeddings is provided. Overall, Dorylus achieves the best performance per dollar against its baseline methods.

$\mathbf{P^3}$~\cite{gandhi2021p3} is a distributed system with sampling-based aggregation focusing on the scenario where a mini-batch can not fit into the memory in a single GPU. Since the main contribution of $P^3$ is a novel communication pattern among workers, we place it in the category of on-device systems. Specifically, $P^3$ assumes that a GNN model's hidden embedding dimension is significantly smaller than that of the input features. Based on this assumption, $P^3$ leverages inter-layer model parallelism for the first layer (i.e., distributedly computing $\mX\mW^{(1)}$ by splitting $\mX$ along the column dimension) to distribute the burdensome storage of input features and then uses data parallelism (i.e., distributively storing $\mA$ for computing node embeddings) for reducing the required communications for transferring intermediate embeddings. Additionally, $P^3$ further pipelines communication and computation across mini-batches for hiding the communication overhead. The limitation of this design is that $P^3$ degrades the performance when the hidden dimension is increased.

\textbf{LLCG}~\cite{ramezani2022learn} is the acronym of ``learn locally correct globally'' which reduces the communication frequency of model synchronization through periodic data transfer. In particular, each training round of LLCG consists of two phases. In the first phase, each worker constructs local mini-batches and updates local parameters following conventional sampling-based training without data transfer. For the second phase, the model parameters in all workers are averaged in a server that keeps on updating the model through mini-batch training with full-graph aggregation. As a variant of partition parallelism, LLCG does not transfer node embedding but synchronizes model parameters instead. The theoretical analysis shows that the stochastic gradient computed by LLCG is bounded by a constant that tends to zero when increasing the size of the sampling size. Since the achievable accuracy of LLCG mainly benefits from the full-graph aggregation on the server side, we regard it as a type of full-graph aggregation method.

\textbf{PipeGCN}~\cite{wan2022pipegcn} optimizes the straightforward implementation of partition parallelism through pipelining embedding computation and the dependent communication across training iterations. In particular, for distributed full-batch training, each worker needs to wait for dependent neighbors' embedding (or embedding gradient) to be transferred before performing local forward (or backward) computation. In PipeGCN, each worker no longer needs to wait for the dependent neighbors. Instead, they use the transferred embeddings/embedding gradients in the previous iteration. This design breaks this computation-communication dependency and allows the training system to perform computation and communication in parallel. Although the model update does not follow traditional gradient descent because it introduces stale embeddings and stale embedding gradients, the convergence analysis shows that PipeGCN enjoys the convergence rate of $\mathcal{O}(T^{-\frac{2}{3}})$, with $T$ being the number of training iterations, which is better than popular sampling-based methods ($\mathcal{O}(T^{-\frac{1}{2}})$). PipeGCN further employs embedding/embedding gradient momentum for mitigating the error caused by the stale data transfer.

\textbf{BNS-GCN}~\cite{wan2022bns} accelerates partition parallelism from a different angle. The authors observe three major issues associated with partition parallelism: significant communication overhead, scaled-out memory requirement, and imbalanced memory distribution. They further observe that both communication volume and memory consumption are correlated to the number of boundary nodes (i.e., the dependent neighbors that need to be received for each partition). To mitigate the issues caused by boundary nodes, BNS-GCN simply adopts Boundary Node Sampling, achieving drastically saved communication, reduced memory, and balanced computation, while maintaining full-graph accuracy. The sampling rate can be as low as 0.1 as suggested by the authors.

\textbf{SAR}~\cite{mostafa2022sequential} also discovers the exploded memory issue in partition parallelism and proposes Sequential Aggregation and Rematerialization scheme. Specifically, each worker receives dependent neighbors through point-to-point communication sequentially. Once the dependent data from one remote worker is received, the receiver worker consumes it, deletes the data to release the memory, and receives the node embeddings from the next worker. For supporting backward propagation, SAR uses reversible neural networks to efficiently reconstruct (rematerialize) the deleted data.

\textbf{Sancus}~\cite{peng2022sancus} improves CAGNET by performing expensive broadcast communication selectively. To avoid overwhelming broadcast communication, which may occupy over 80\% of the total training time, each worker maintains a copy of other workers' embeddings and only refreshes them when the embeddings are too stale. Based on this design, Sancus is able to skip considerable rounds of communication, thus achieving better throughput. Three staleness metrics are proposed for the staleness check.

\subsection{Swap-based scalable training systems}

\textbf{GNNAdvisor}~\citep{wang2021gnnadvisor} is another work that targets scalable GNN training with limited devices. Compared with NeuGraph and Roc, GNNAdvisor further explores feature-level partition. It implements a low-overhead cost model to determine the best graph-level and feature-level partition. The current implementation of GNNAdvisor only supports single-device training.

\textbf{GNNAutoScale}~\citep{fey2021gnnautoscale} and \textbf{GraphFM}~\citep{yu2022graphfm} develop scalable algorithms for efficient GNN systems. Specifically, to enable the training within limited GPU memory, GNNAutoScale leverages ClusterGCN (described in Section~\ref{subsec:sampling}). During the training process, all nodes maintain their features and the latest intermediate embedding in the host CPU memory. For better approximating neighbor aggregation during the training, GNNAutoScale accesses all maintained historical embeddings for unsampled neighbors. With this design, all nodes can access all neighbor embeddings although the embeddings of unsampled nodes are stale. GraphFM is a follow-up work of GNNAutoScale that develops GraphFM-IB and improves the approximated neighbor aggregation by integrating feature momentum for alleviating incurred staleness. Their convergence analysis shows that GraphFM-IB does not rely on large sampling size. To further mitigate staleness, a variant named GraphFM-OM is developed by pushing the intermediate embeddings of sampled nodes to update the unsampled nodes, achieving better accuracies.

\subsection{Sampling-based scalable training systems}

\textbf{PaGraph}~\cite{lin2020pagraph} is a multi-GPU system for GNN training. For reducing the data transfer between CPU and GPU for neighbor sampling, PaGraph heuristically minimizes data movement by caching nodes with large out-degrees in GPUs. To enable multi-GPU training, PaGraph devices a graph partition algorithm to balance training nodes across partitions, achieving approximately balanced workload. Each partition only maintains nodes within the partition and $L$-hop neighbors for an $L$-layer model for avoiding redundant storage.

\textbf{DistDGL}~\cite{zheng2020distdgl} and \textbf{DistDGLv2}~\cite{zheng2021distributed} are distributed training frameworks based on DGL~\cite{wang2019dgl}. DistDGL is developed for CPU clusters where each worker maintains a partitioned subgraph in the host memory with its node features and one-hop neighbors. During the training process, each worker first launches processes for mini-batch sampling which create training batches while fetching features of sampled remote neighbors. METIS is used for balancing training nodes across partitions and reducing edge cuts so that the communication across workers is minimized. To further approximately balance workload, multi-constraint partition is adopted for balancing nodes and edges. For the graphs with learnable node embedding, DistDGL adopts asynchronous update of sparse embedding for overlapping communication and computation while enjoying seldom data access conflicts. DistDGLv2 extends DistDGL to heterogeneous graphs and GPU clusters and deploys asynchronous mini-batch generation for fully utilizing all computation resources. Overall, DistDGLv2 is 2$\times$-3$\times$ faster than DistDGL. Both projects have be integrated with DGL and DistDGLv2's APIs are compatible with DGL.




\textbf{SALIENT}~\cite{kaler2022accelerating} boosts the training of GNNs with an efficient dataloader. This work identifies that for the conventional implementation of GNN training where neighbor sampling is conducted in CPU and the GNN model is computed in GPUs, batch preparation and data transfer takes around 72\% total epoch time, which severely degrades the efficiency of GNN training. In particular, batch preparation contains two stages: 1) sampling and creating message-flow graphs; and 2) fetching feature and labels through tensor slicing. The first stage has a large optimization space because neighbor sampling can be accomplished by multiple design choices and implementation (e.g., mapping node indices between the original graph and the sampled message-flow graph, fusing the operations of neighbor sampling and minibatch construction). SALIENT empirically explores a fast sampler that is microbenchmarked in ogbn-products \cite{ogb} for each individual hop, which yields 2.5x faster speed than PyG. For accelerating the second stage, SALIENT implements C++ threads for parallelly slicing tensors in shared memory. To further mitigate the overhead of data transfer, SALIENT pipelines data transfer and GPU computation. SALIENT also evaluates neighbor sampling for efficient GNN inference.

\textbf{GNNLab}~\cite{yang2022gnnlab} improves PaGraph with sampling minibatches in GPUs. This improvement is not trivial because on-device sampling requires extra memory which restricts cache size thus increases missing rates. In addition, simply caching nodes with large out-degree is not optimal because it ignores the pattern of training nodes. Therefore, some highly-connected nodes are not frequently sampled during the training process. To address this issue, GNNLab first performs serveral rounds of pre-sampling. Most frequently visited nodes are selected as cached nodes. In addition to the pre-sampling based caching policy, GNNLab further enhances cache size through space-sharing design. Some GPUs are selected as samplers and the others are trainers. This strategy enhances data locality for all workers. To balance the execution time of samplers and trainers, GNNLab first determines the number of samplers and trainers based on their execution time in one epoch so that samplers can finishes sampling tasks in shorter time, and then, idle samplers are switched to trainers for fully leverage computation resources. 

\textbf{BGL}~\cite{liu2021bgl} also observes data movement bottlenecks in sampling-based GNN training systems and develops a dynamic GPU cache by leveraging temporal locality among mini-batches, which can be strengthened by selecting training nodes via BFS sequences. However, a straightforward implementation incurs biased label distribution. To mitigate this issue, BGL generates multiple BFS sequences by randomly selecting BFS roots. Furthermore, for each training epoch, the BFS sequences will be shifted to add randomness for batch generation. Under the multi-GPU setting, BGL ensures no duplicated cached nodes because cached nodes can be shared via NVLink and this design saves GPU memory. BGL further develops a scalable graph partition algorithm with BFS while heuristically ensuring a balanced workload. Finally, BGL optimizes resource allocation of data preprocessing by formulating it as an optimization problem.


\textbf{GNS}~\cite{dong2021global} stands for Global Neighbor Sampling, which develops an efficient and scalable sampling algorithm for reducing the data movement between CPU and GPU. To achieve this, GNS periodically select a subset of all nodes and caching their node features in GPUs. During the neighbor sampling process, if the GPU caches contain sufficient neighbors, GNS only samples neighbors from the caches. Otherwise, extra neighbors outside GPUs are sampled. GNS devices two caching strategies: 1) the nodes are cached with the probability proportional to their degrees so that only a small cache needs to be maintained for a power-law graph; 2) the sampling probability for each node is determined by random walks where more frequently visited nodes are more likely to be sampled.

\textbf{MariusGNN}~\citep{waleffe2022marius++} is a GNN training system targeting deeper vertical scalability, which is developed based on Marius \cite{273733}. The above-mentioned works assume that all data can be saved in the host CPU memory. This work considers a more resource-constraint scenario where CPU memory is limited. For reducing expensive I/O operations, MariusGNN develops a sophisticated dataloader, which enjoys reduced I/O operations and continuous external memory access.

\textbf{DSP}~\cite{cai2023dsp} accelerates GNNs by addressing two main challenges associated with multi-GPU training: low GPU utilization and high communication costs. To address the first challenge, DSP uses a tailored data layout that stores the graph topology and popular node features in GPU memory and leverages NVLink to share GPU caches among all trainers, which allows for efficient graph sampling with multiple GPUs. Additionally, DSP introduces a collective sampling primitive (CSP) that pushes the sampling tasks to data to reduce communication costs. To address the second challenge, DSP employs a producer-consumer-based pipeline that allows tasks from different mini-batches to run congruently, improving GPU utilization.

\section{Detailed Discussion of Customized Hardware} \label{app:hardware}

\subsection{General Workloads:All-Stage Unified Architecture}

\textbf{EnGN} \cite{liang2020engn} targets graph convolution network (GCN)~\cite{gcn}, relational GCN (R-GCN)~\cite{rgcn2018}, gated GCN~\cite{gatedgcn2017}, GraphSage~\cite{sage}, and graph recurrent network~\cite{grn2017}. To process them, the operations are divided into three stages: feature extraction, aggregation, and update. The authors develop a Neural Graph Processing Unit (NGPU) as a unified architecture for executing them. These stages are processed as pipelined matrix multiplications of the feature matrix, neural weights, and adjacency matrix. The NGPU uses a 32-bit fixed-point $128 \times 16$ systolic array as the main computation unit, which maps nodes to different rows and their features to different columns. To perform the aggregation, the authors propose a ring-edge-reduce dataflow that connects processing elements (PEs) in the same column with a ring so that the results can be passed through and added together based on the adjacency matrix. As the adjacency matrix is sparse, the authors reorganize the edges to reduce the number of idle PEs during the execution of aggregation. Additionally, high-degree nodes are cached to decrease the number of off-chip transactions since they are reused across multiple operations. 
EnGN analyzes the effect of different schedules on the number of operations and develops a dimension-aware stage reordering strategy, which changes the order of execution between the aggregation and feature extraction stage based on the input and output dimensions, to reduce the number of operations. Furthermore, it splits the graphs into 2D tiles to fit them into on-chip resources and develops an analytical formula to decide whether for a given layer, a column-order traversal of the tiles results in fewer I/O operations or their row-order traversal.

\textbf{Rubik}~\cite{chen2021rubik} develops  an accelerator tailored to train GIN and GraphSage. To enhance graph locality and data reuse, the proposed design leverages a pre-processor that reorders the graph to group nodes with common neighbors. The graph subsequently passes through a hierarchical accelerator architecture, consisting of a PE array interconnected via a 2D-mesh network NoC. Each PE embedded in this architecture features a multiply-and-accumulate (MAC) array. To map the graph and features to the accelerator, the authors presented graph-level and node-level mappings. The former entails mapping a window of nodes to one PE, allowing for task-level parallelism, with PEs working on distinct node windows. The latter involves tiling the dense vector-matrix multiplication for the feature update (aggregation and update) using the MAC array.
Rubik incorporates a global buffer for the PE array, two private caches (for data and instruction) in each PE, and register files (RFs) in every MAC to facilitate data reuse.

\textbf{G-CoS} \cite{zhang2021g} is an automated framework that aims to optimize the performance and efficiency of GNNs by co-searching for the best GNN structures and accelerators. This framework consists of two components: (1) a one-shot co-search algorithm for GNN structures and their matched accelerators, and (2) a generic accelerator search space that can be applied to various GNN structures. The accelerator search space in G-CoS is based on a template comprising multiple sub-accelerators that can handle both sparse and dense matrix multiplications. By configuring the settings of each sub-accelerator, such as tiling sizes, tiling order, and interconnection style, they can be specialized to process different clusters of GNN operators with distinct data sizes and sparsity, which ultimately enhances the hardware efficiency. In addition, G-CoS employs local buffers assigned to the intermediate features, index for sparse features (in COO format), and weights and intermediate outputs for each sub-accelerator. These buffers can be interconnected to allow for data sharing and reuse across different sub-accelerators, reducing the need for costly off-chip memory access. G-CoS is a pioneering effort towards automating the search for both GNN model structure and accelerator design knobs.

\textbf{GCoD} \cite{you2022gcod} is a co-design framework that addresses the challenge of extreme sparsity in GNN inference by optimizing both the algorithm and hardware accelerator. At the algorithm level, GCoD polarizes the graphs into either denser or sparser local neighborhoods without compromising model accuracy. This polarization results in adjacency matrices with two levels of workload, which greatly improves regularity and ease of acceleration. On the hardware level, GCoD integrates a dedicated two-pronged accelerator comprising a dense branch and a sparse branch processor. The dense branch processor employs a chunk-based micro-architecture with several heterogeneous computing modules to balance the workload across different subgraphs from the polarized adjacency matrix. This balance is achieved by allocating computing resources and bandwidth proportionally based on the data volume and operation size involved in each subgraph's processing. The sparse branch processor accelerates the remaining sparse workloads, which constitute a small portion of nonzero data, mostly on-chip. To store data in a sparse format, the sparse branch processor uses the CSC format, which significantly reduces the storage overhead of adjacency matrices. GCoD's sparse branch processor also employs a query-based weight forwarding mechanism that flexibly shares on-chip input data from the dense branch processor to enhance data reuse. The weight forwarding is performed on-demand of the sparse branch to achieve more efficient control, and it accesses around 63\% of the data, which drastically reduces the off-chip access cost. Overall, GCoD's algorithm-hardware co-design framework offers better accuracy-efficiency trade-offs than optimizing each aspect separately. By polarizing the graphs into denser and sparser local neighborhoods and integrating a two-pronged accelerator with a dense and sparse branch processor, GCoD provides a comprehensive solution to tackle the challenge of extreme sparsity in GNN inference.

\textbf{I-GCN} \cite{geng2021gcn} proposes a new approach called "islandization" to enhance data locality in GNNs. By identifying clusters of nodes with strong internal connections, called "islands," I-GCN aims to minimize off-chip memory accesses and improve on-chip data reuse. To support islandization, I-GCN's hardware architecture includes an Island Locator and an Island Consumer. The Island Locator searches the graph from multiple nodes in parallel to locate highly-connected nodes termed as hubs. From these hubs, multiple tailored breadth-first-search engines are used to identify islands. The Island Consumer then combines and aggregates the islands and hubs, with a focus on reusing aggregation results among nodes with shared neighbors. To optimize the aggregation process, a ring-based on-chip network is used to distribute partial results and detect reuse opportunities among processing elements (PEs), and an in-network reduction scheme is employed to minimize network communication latency. Overall, I-GCN's islandization approach and hardware design achieve better data locality and efficiency in GNNs.

\subsection{General Workloads: Dedicated Module per Stage}

\textbf{HyGCN}~\cite{yan2020hygcn} targets GCN, GraphSage, GINConv~\cite{ginconv2018}, and DiffPool~\cite{diffpool} and treats their operations as two main stages, aggregation, and combination (Update in Eq.~\ref{eq:message_passing}). Due to the distinctive execution patterns of the aggregation and combination stages, the proposed accelerator employs a separate processing engine for each stage and processes them in a dataflow manner. Specifically, the Aggregation engine utilizes parallel single-instruction multiple-data (SIMD) cores to exploit the intra-vertex parallelism resulting from the long feature vectors assigned to each vertex by the GNN algorithms. If the feature vector length is smaller than the total number of cores, the unassigned cores are allocated to other nodes, thereby enabling edge-level parallelism for this stage. As the graph connections are sparsely distributed, a sparsity eliminator is employed to identify a region in the adjacency matrix of the corresponding graph partition that stores the effective edges, using a window sliding and shrinkage technique.
To further reduce the computation complexity, the aggregation operation is performed on a sampled set of neighbors. The sparsity eliminator and sampler also help to avoid fetching feature vectors of the nodes that are not connected to any edge in the current graph partition.
 Subsequently, the Combination engine treats the computation as a dense matrix-vector multiplication (MVM) and is realized as a group of systolic arrays that can work independently or cooperatively depending on the configuration. Each systolic array processes a small group of nodes, with the feature vectors (weights) of the nodes flowing through the row (column) dimension.

\textbf{FlowGNN} \cite{sarkar2022flowgnn} proposes a generic dataflow architecture for accelerating GNNs that can flexibly support the majority of message-passing GNN algorithms, including GCN, GINConv, PNA~\cite{pna2020}, GAT~\cite{gat}, DGN~\cite{beaini2021directional}, and GNN with virtual node (VN)~\cite{mpnn}. The authors argued that common pre-processing techniques used to exploit data locality are not feasible for real-time applications with millions of input graphs with varied structures. Thus, they directly take the graph in COO format without any pre-processing to reorganize the data. FlowGNN's architecture is based on the idea that any graph-related functions can be expressed with pair-wise message passing. It divides the operations into three steps: 1) the gather phase, which aggregates messages from neighbors; 2) the node transformation (NT) phase, which applies the update function (or feature transformation function) to messages; and 3) the scatter phase, which constructs messages for the next layer by applying the respective message transformation function (e.g., applying the edge weights) on the node embeddings produced in step two. The computations are divided into these steps to integrate edge embeddings and different aggregation functions. The gather and scatter phases handle per-edge computations and are implemented as a single unit (MP unit) that follows the NT unit in a dataflow fashion. The NT unit handles per-node computations. 
As such, the MP (NT) unit consists of parallel PEs that parallelize the edges (nodes). Each of the PEs further parallelizes the features. To avoid bank conflicts, each of the PEs in the MP unit processes one particular bank of edges. FlowGNN uses an NT-to-MP adapter to perform on-the-fly multicasting of the node embeddings generated by the NT unit to their right PE in the MP unit.

\textbf{BlockGNN}~\cite{zhou2021blockgnn} aims to compress the update stage (feature transformation) of GNN models for GraphSage-Pool (GS-Pool), GCN, GGCN~\cite{ggcn}, and GAT that require a large weight memory, high computing resources, and high computing latency when the feature size is large. To achieve this, they leverage block-circulant weight matrices, which are configured in the form of block-circulant matrices that can be accelerated by Fast Fourier Transform (FFT) and Inverse Fast Fourier Transform (IFFT). This transformation reduces the computation intensity and required storage for the update stage. 
The computation engine of BlockGNN is composed of two main units, the CirCore unit for the update stage and the VPU (vector processing unit) for the aggregation stage and non-linear operations. The CirCore unit is a three-stage pipeline, consisting of an FFT unit that converts data from the spatial domain to the spectral domain, a systolic array that performs element-wise product and accumulation, and an IFFT unit that transforms the results back to the spatial domain. The VPU is organized as a SIMD unit with m lanes, each with a parallel factor of 16.

\textbf{DeepBurning-GL}~\cite{liang2020deepburning} is an automated framework designed to facilitate the development of hardware accelerators for GNNs. The framework takes a GNN model as input, which is designed using software frameworks such as DGL and PyG, and generates a hardware accelerator for the target FPGA platform. For the hardware design, the authors provide three templates. The first template is the GNN computation template that handles dense and sparse multiplications. Dense multiplications are implemented via a systolic array or a dot production array, while sparse multiplications are handled by an array of SIMD units or a ring-reduce topology (similar to EnGN). The second template is the memory template, which consists of a distributed on-chip buffer, cache, or degree-aware hierarchy on-chip buffer to support both regular and irregular memory accesses. The third template is the graph manipulation template, which supports graph sampling and reconstruction in various GNN models. To adapt to the specific characteristics of the target model and FPGA platform, the authors develop analytical models that analyze the processing requirements and adopt a design space exploration to tune hardware parameters for the required templates based on the graph properties, network architecture, and resource constraints. The authors validate their framework on various GNN algorithms, including GCN, GS-Pool, R-GCN, and EdgeConv~\cite{wang2019dynamic}.

\subsubsection{Special Workload: Layer-Customizable Deep Pipeline.} 

\textbf{AWB-GCN} \cite{geng2020awb} argues that many real-world graphs follow the power-law distribution, which implies that the number of nodes with a given degree x follows a proportional relationship with $x^{-\beta}$ for a constant $\beta$. This leads to a situation where a small number of rows/columns in the adjacency matrix contain the majority of non-zero elements, while the rest contain only a few non-zero elements. The authors also note that ReLU activation leads to an abundant number of zero elements, making it more efficient to treat the feature transformation step (Update step) as a sparse computation.
In response to these observations, the authors propose an accelerator architecture called Autotuning-Workload-Balancing (AWB) GCN, which includes \textit{distribution smoothing}, \textit{evil row remapping}, and \textit{remote switching}. 
Since GCNs have linear aggregation and update functions, the authors propose executing the update stage before the aggregation stage since this approach helps reduce the number of non-zero operations. This is because by this change, in both steps, we are dealing with two SpMM as opposed to an SpMM and a dense matrix multiplication.
AWB-GCN's architecture supports inter-layer parallelism in addition to the commonly used intra-layer parallelism and is based on the inner-product matrix multiplication. The authors partition the graph and implement a task distributor to navigate the data to idle PEs to balance the workload. The architectural optimizations are designed to overcome the load imbalance problem. 
Distribution smoothing is local and tries to balance the load among neighboring PEs within up to three hops. The remote switching addresses regional clustering by exchanging the workloads of under- and overloaded PEs. Evil row remapping distributes rows with the most non-zeros to the most under-loaded PEs.

\textbf{StreamGCN} \cite{sohrabizadeh2022streamgcn} presents an efficient and flexible GCN accelerator for streaming small graphs - from the DRAM, the host CPU, and through the network - and exploiting all the available sparsity. The authors argue that in memory-bound applications like this, it is essential to minimize memory transactions, and they propose a scheduling mechanism to achieve this. In this scheduling, input data is fetched from global memory (DRAM) only once and is reused for all corresponding computations before being evicted. The proposed GCN accelerator includes dedicated modules for each layer, which not only allows customization of parallelization based on the workload of each layer but also enables inter-layer pipelining. As a result, the intermediate results are directly passed to the modules for the next layer through on-chip FIFOs which further helps with decreasing the number of DRAM transactions. To avoid read-after-write (RAW) dependencies in the aggregation unit, the edges are pre-processed and reordered. Both the aggregation unit and the feature transformation unit contain multiple SIMD PE with the SIMD dimension parallelizing the node features. The PEs in the aggregation unit process different pairs of neighbors in parallel with their respective edge weight (edge parallelism), while in the feature transformation unit, they parallelize the nodes. These units are connected in a dataflow fashion.
Similar to AWB-GCN, StreamGCN treats the feature transformation step as a sparse computation. However, due to the smaller size of the target graphs, the method is based on outer-product-based multiplication. This scheduling is shown to reduce the number of pipeline stalls due to RAW dependencies. Since the intermediate results are to directly be processed as they are being produced, StreamGCN has in-situ sparsity support by developing a mechanism to prune the zero elements of the node embeddings on-the-fly, while they are being generated. This is realized by a pruner and an arbiter unit. The pruner pre-fetches the results at a higher rate than the next layer's processing rate to pass non-zero elements through FIFOs. The arbiter then checks them for RAW dependencies and dispatches enough elements to fill the PEs. 
Overall, the proposed StreamGCN architecture is ideal for real-time or near-real-time graph search and similarity computation for many biological, chemical, or pharmaceutical applications.

\subsection{Special Workload: All-Layer Unified Architecture.}
As seen in the previous section, a GCN algorithm consists of multiple layers that have distinct features (e.g., node embedding dimension).
In contrast to the works detailed in Section~\ref{sec:customized_hw_custom_layer}, the studies presented in this category put forth the idea of constructing a more adaptable architecture across distinct layers and utilizing the same engines for all layers. 
Since the works in this category focus on a specific workload, namely GCN, they offer greater possibilities for customization than those presented in Section~\ref{sec:customized_hw_general}, as indicated below:

\textbf{GraphACT} \cite{zeng2020graphact} develops an accelerator for training GCNs targeting small graphs on CPU-FPGA heterogeneous systems. The small graphs are created by sampling subgraphs of the input graph to be able to fit them into the on-chip resources. 
GraphACT does not work directly with the normalized adjacency matrix and rather defines three types of multiplications for GCNs: 1) dense (weights) - dense (embeddings), 2) binary sparse (adjacency matrix) - dense, 3) diagonal (degree matrix) - dense. The first multiplication is performed using a 2D systolic array called the Weight Transformation module, where nodes are mapped to the row dimension and output features to the column dimension. The Feature Aggregation module implements steps two and three. The third multiplication (Step three) is treated as scaling the rows of the second matrix in the multiplication with minimal hardware overhead.
GraphACT implements a redundancy reduction technique to reduce the computation cost of the second step, which seeks to pre-compute the repeated aggregations on CPU. This is possible since GraphACT directly works with the adjacency matrix in its \textit{binary} (not normalized) format. The redundancy reduction algorithm identifies the common pairs of neighbors and merges each of the most repeated ones in a node. This is done as a pre-processing step on CPU since it is communication-intensive.
 The Feature Aggregation module has a 1D accumulator array that parallelizes the feature dimension.  It adds the features of the identified node pairs and performs the aggregation based on the adjacency matrix of the merged graph (generated after merging the node pairs). Finally, it scales the node features based on the node degrees (Step 3). Both the Feature Aggregation and Weight Transformation modules are reused for all the \textit{$L$} layers of the GCN.

\textbf{Zhang et al.}~\cite{zhang2020hardware} focus on accelerating GCN inference for large graphs. The approach begins by partitioning the input data into smaller sizes that can fit into the on-chip resources for processing. Following this, a redundancy reduction scheme, such as GraphACT, and a node-reordering phase are applied to decrease global memory access. Upon completion of the pre-processing stage, the graph is passed to a hardware architecture that incorporates two primary modules: Aggregation and Weight Transformation. 
The Aggregation module employs parallel vector accumulators (VAs) for sparse matrix multiplication in the GCN aggregation process. Here, each VA parallelizes the feature dimension and different VAs process distinct edges. The Weight Transformation module leverages a systolic array to implement the update stage. These modules are bi-directionally connected to enable different modes of scheduling, where the order of executing the aggregation and update stage can vary based on the GCN and dataset properties.

\textbf{BoostGCN}~\cite{zhang2021boostgcn} presents a framework aimed at optimizing GCN inference on FPGAs. The hardware architecture of BoostGCN includes two primary modules, namely the feature aggregation module (FAM) and the feature update module (FUM). These modules are interconnected through internal buffers that cache intermediate results, and the connection between them is bi-directional, as in Zhang et al.'s work. To reduce memory traffic and address workload imbalance, the authors propose Partition-Centric Feature Aggregation (PCFA), which leverages tiling in three dimensions: edges, nodes, and features. FAM operates on multiple edges in parallel and further parallelizes the feature dimension for them. Additionally, FAM uses a sort-and-combine (SaC) unit to sort the inputs and combine them if two of them have the same destination, which reduces congestion. BoostGCN employs two types of FUM based on the sparsity of the feature matrix (node embeddings) -- Sparse-FUM and Dense-FUM. Sparse-FUM uses the same structure as FAM, with a format transformation module in the beginning that transforms the input to COO format. Dense-FUM, on the other hand, implements a 2D systolic array.

\section{Detailed Discussion of Special Graphs and Special GNNs} \label{app:broader}

\subsection{Heterogeneous Graphs}

\textbf{HetGNN}~\cite{hetgnn} is a GNN on heterogeneous graphs that introduces a novel heterogeneous neighbor sampler. The sampler belongs to the node-wise sampling category discussed in Section~\ref{subsubsec:node-wise} but overcomes a special challenge on heterogeneous graphs. On heterogeneous graphs, node types in a target node's receptive field can be very imbalanced, e.g., some types are very infrequent compared to other types. HetGNN argues that a regular sampler may miss some less common node types completely and result in insufficient node representations. Therefore, instead of randomly sampling, the HetGNN sampler first runs random walks with restart (RWR) in the neighborhood and then chooses nodes to form the computation graph according to their hit frequency by RWR. The sampler will also balance node types, such that the top frequent nodes of all types are chosen. With this RWR-based and type-balanced sampler, HetGNN can achieve fast training with good accuracy performance.

\textbf{HGT}~\cite{hu2020heterogeneous} is a transformer-based GNN on heterogeneous graphs, which also introduces an efficient sampler to make the model handle Web-scale graph data with billions of edges. The HGT samples subgraphs in mini-batches. The sampler is able to 1) maintain a similar number of nodes and edges for each type in each subgraph and 2) keep the sampled subgraph dense. Specifically, for each node type it maintains a node budget and iteratively samples the same number of nodes with an importance sampling strategy to reduce variance. The sampling probability for each node type is calculated based on the square of the cumulative normalized degree of other nodes that have already been sampled in its budget. The model is shown to achieve state-of-the-art results on the Open Academic Graph (OAG) which contains millions of nodes and billions of edges.

\textbf{Pigeon}~\cite{wu2023pigeon} is an intermediate representation (IR) and code generator for end-to-end training and inference of relational graph neural networks (RGNNs)~\cite{rgcn2018} on heterogeneous graphs. Pigeon addresses the performance challenges posed by RGNNs by decoupling the model semantics, data layout, and operator-specific schedule, and expressing these opportunities to allow them to be integrated into the design space as integral elements. The proposed solution includes a compact tensor materialization scheme and a linear operator fusion pass to achieve both training and inference. The compact tensor materialization scheme reduces the resources spent on computing and storing common subexpressions in RGNNs. Specifically, certain edge data are determined by source node features and edge types, and instead of computing and storing such data for each edge, the system computes and stores the data once for each (edge type, unique node index) pair. This reduces the amount of computation and storage required for common subexpressions. The linear operator fusion pass switches the order of linear operators, such as linear layers and inner products. When a linear operator is followed by another linear operator, their order may be switched to reduce the strength.

\subsection{Dynamic Graphs}

\textbf{TGL} ~\cite{zhou2022tgl} proposes a unified framework for accelerating continuous-time temporal GNN training. It designs a temporal-CSR data structure for rapid access of temporal neighborhood candidates, and a parallel sampler is further equipped to accelerate temporal neighborhood sampling. Also, it proposes a novel random chunk scheduling technique for training acceleration where simply increasing the batch size would result in the obsolete node memory issue. TGL ~\cite{zhou2022tgl} shows its great capability in handling large-scale dynamic graphs where it achieves an average of 173$\times$ speedup on a multi-core CPU compared with the baselines.

\textbf{MTGNN}~\cite{wu2020MTGNN} applies GNNs to model the spatio-temporal graph for solving the multivariate time series prediction problem, where variables are dependent on each other but their dependence (graph structure) is latent. Naively assuming a fully-connected graph structure will result in $N^2$ edges and impose huge GNN computation time. MTGNN proposes a graph learning module to automatically extract the uni-directed relations among nodes, which greatly simplifies the structure and accelerates GNN computation. MTGNN also applies clustering to reduce memory usage by the intermediate states of nodes and save memory access time. Specifically, the edge search space is reduced by first randomly separating nodes into clusters and then only forming edges within each cluster. The clustering process for nodes is repeated in each iteration, which allows every pair of nodes in the same cluster to be possibly connected.

\textbf{GraphODEs}~\cite{zijie2020lgode, zijie2021cgode} combine the expressive power of GNNs with the principled ordinary differential equation (ODE) modeling for learning dynamical systems. \textbf{LG-ODE}~\cite{zijie2020lgode} is a latent ODE generative model, which models multi-agent system dynamics with fixed graph structure in a continuous manner. The ODE function in LG-ODE is a GNN to capture the continuous interaction among agents, and the latent initial states for ODEs are inferred through another GNN-based encoder. Towards acceleration, existing discrete neural models learn a fixed-step transition function that takes the system state at time $t$ as input to predict the state at time $t+1$. An adaptive ODE solver will automatically adjust the times, where the graph interaction module is called to provide the balance between model performance and time consumption. Additionally, LG-ODE is especially superior at making long-range predictions. It is able to adjust its prediction length without re-feeding its previous predictions at the same fixed step during training, which greatly accelerates model inference. Moreover, since LG-ODE is only capable of handling dynamic graphs with evolving nodes but fixed edges, the follow-up work \textbf{CG-ODE}~\cite{zijie2021cgode} is proposed to jointly model the evolution of nodes and edges through two coupled ODE functions. 

\textbf{GN-ETA}~\cite{derrow2021eta} proposes a large-scale GNN model for estimating the time of arrival (ETA) in real time. Supersegment-based acceleration is applied so that the complex interaction of traffic roads and the traffic conditions evolution can be joint modeled over time. In particular, GN-ETA segments the traffic network into several supersegments and is trained to predict both the across-supersegments time and within-supersegments time. For model inference, the sequence of supersegments will be put in a route sequentially where the later supersegements would utilize the prediction time of an earlier one to determine its relevant prediction horizons for scaling up. Additionally, GN-ETA adopts the MetaGradients methodology~\cite{xu2018meta} to stabilize the training process raised by uneven query batches that are commonly seen in real-life applications. GN-ETA is able to handle web-scale applications like Google Maps. Recently, the follow-up work CompactETA~\cite{fu2020eta} proposes to encode temporal information using positional encoding instead of recurrent networks to further accelerate the model.

\end{document}